\documentclass[lettersize,journal]{IEEEtran}

\usepackage{amsmath,amsfonts}
\usepackage{algorithmic}
\usepackage{algorithm}
\usepackage{array}
\usepackage{pifont}
\usepackage[caption=false,font=normalsize,labelfont=sf,textfont=sf]{subfig}
\usepackage{textcomp}
\usepackage{stfloats}
\usepackage{url}
\usepackage{verbatim}
\usepackage{graphicx}
\usepackage{cite}
\usepackage{comment}
\usepackage[colorlinks]{hyperref}
\usepackage{booktabs}
\usepackage{enumerate}
\usepackage{enumitem}
\usepackage{makecell}
\usepackage{tabu}
\usepackage{multirow}
\usepackage{utfsym}
\usepackage{colortbl} 
\usepackage{soul}
\definecolor{mygray}{gray}{.9}

\newcommand{\ie}{\textit{i}.\textit{e}.}
\newcommand{\eg}{\textit{e}.\textit{g}.}

\begin{document}

\title{Towards Unified Referring Expression Segmentation Across Omni-Level Visual Target Granularities}

\author{Jing Liu*\textsuperscript{\dag}, \IEEEmembership{Member, IEEE}, Wenxuan Wang*, Yisi Zhang, Yepeng Tang, Xingjian He, Longteng Guo, Tongtian Yue, Xinlong Wang
\thanks{
This research is supported by Artificial Intelligence-National Science and Technology Major Project (2023ZD0121200) and the National Natural Science Foundation of China (6243000159, 62102416), and the Key Research and Development Program of Jiangsu Province under Grant BE2023016-3. 
}
\thanks{J. Liu, W. Wang, X. He, L. Guo and T. Yue are with the Institute of Automation, Chinese Academy of Sciences, Beijing 100190, China, while J. Liu and W. Wang are also with the School of Artificial Intelligence, University of Chinese Academy of Science, Beijing 100190, China. (e-mail: jliu@nlpr.ia.ac.cn, wangwenxuan2023@ia.ac.cn)}
\thanks{W. Wang and X. Wang are with the Beijing Academy of Artificial Intelligence, Beijing 100085, China. (e-mail: wangxinlong@baai.ac.cn)}
\thanks{Y. Zhang is with the University of Science and Technology Beijing, Beijing 100083, China.}
\thanks{Y. Tang is with the School of Computer Science and Technology, Beijing Jiaotong University, Beijing 100044, China.}
\thanks{* Equal Contribution. \textsuperscript{\dag}Corresponding author.}
}

\maketitle

\begin{abstract}

Referring expression segmentation (RES) aims at segmenting the entities' masks that match the descriptive language expression. While traditional RES methods primarily address object-level grounding, real-world scenarios demand a more versatile framework that can handle multiple levels of target granularity, such as multi-object, single object or part-level references. This introduces great challenges due to the diverse and nuanced ways users describe targets.
However, existing datasets and models mainly focus on designing grounding specialists for object-level target localization, lacking the necessary data resources and unified frameworks for the more practical multi-grained RES.
In this paper, we take a step further towards visual granularity unified RES task.
To overcome the limitation of data scarcity, we introduce a new multi-granularity referring expression segmentation (MRES) task, alongside the RefCOCOm benchmark, which includes part-level annotations for advancing finer-grained visual understanding.
In addition, we create MRES-32M, the largest visual grounding dataset, comprising over 32.2M masks and captions across 1M images, specifically designed for part-level vision-language grounding. 
To tackle the challenges of multi-granularity RES, we propose UniRES++, a unified multimodal large language model that integrates object-level and part-level RES tasks. UniRES++ incorporates targeted designs for fine-grained visual feature exploration. 
With the joint model architecture and parameters, UniRES++ achieves state-of-the-art performance across multiple benchmarks, including RefCOCOm for MRES, gRefCOCO for generalized RES, and RefCOCO, RefCOCO+, RefCOCOg for classic RES. 
To foster future research into multi-grained visual grounding, our RefCOCOm benchmark, MRES-32M dataset and model UniRES++ will be publicly available at \textit {\url{https://github.com/Rubics-Xuan/MRES}}.

\end{abstract}


\section{Introduction}
\label{introduction}

\IEEEPARstart{R}{esearch} community has recently observed significant progress in multimodal embodied intelligence \cite{ahn2022can,reed2022generalist,driess2023palm,shah2023lm,gao2023physically,brohan2023rt}. 
Within this broader landscape, referring expression segmentation (RES) has emerged as a critical visual grounding task, demanding the embodied agents to precisely pinpoint the target objects at pixel level based on natural language descriptions. 
Unlike traditional visual segmentation tasks that operate within predefined categories, RES must contend with the open-world nature of real-world scenes, where regions of interest are highly diverse and context-dependent. 
This flexibility underscores the importance of RES in scenarios where fine-grained localization is necessary. 
Notably, user-provided expressions often vary in region specificity, reflecting grounding needs across diverse levels of target granularity (\ie, multiple objects, a single object, or even a part of an object at a time), which does not typically point to a single object region as in classic RES task. 
This multi-grained variability in practical scenes not only poses greater challenges for RES systems, requiring accurate interpretation and segmentation of targets across varying levels of detail, but also highlights the crucial role of RES in enabling more adaptive and context-aware vision-language (V-L) interactions. 
By pushing traditional RES focusing solely on object-level categories towards systems that handle a wide spectrum of target granularity, multi-grained RES can foster more intuitive and effective human-machine interactions, which closely aligns with the nuanced ways users express their needs in real-world settings.

\begin{table}[t]
  \renewcommand\arraystretch{1.2}
  \centering
  \footnotesize
  \caption{Comparison among different referring expression datasets, including ReferIt \cite{kazemzadeh2014referitgame}, RefCOCO(+/g) \cite{yu2016modeling,nagaraja2016modeling}, PhraseCut \cite{wu2020phrasecut}, and our proposed \textbf{MRES-32M}. Part-Level: expressions that specifies various parts of the target object in the given image.}
  \vspace{-3mm}
  \setlength{\tabcolsep}{0.6mm}
  {\begin{tabular}{lcccc}
    \specialrule{.1em}{.05em}{.05em} 
          Property & ReferIt & RefCOCO(+/g) & PhraseCut&  \textbf{MRES-32M}\\
          \cline{1-5}
          Image Source  & CLEF \cite{grubinger2006iapr} & COCO \cite{lin2014microsoft} & VG \cite{krishna2017visual}  & Object365 \cite{shao2019objects365} \\
          Object-Level & \ding{51} & \ding{51} & \ding{51} & \ding{51} \\
          Part-Level & \ding{53} & \ding{53} & \ding{53} & \ding{51} \\
          \makecell[c]{Expression Type} & Free  & Free  & Templated & Free \\
    \specialrule{.1em}{.05em}{.05em} 
    \end{tabular}}
  \label{tab:dataset_compare}
\vspace{-15pt}
\end{table}

Contrary to the classic RES task, realizing the more practical multi-grained RES involves two vital aspects that require consideration.
1) \emph{Multi-Granularity Grounding Data Scarcity}: 
Since the concept of the RES task is initially proposed in \cite{hu2016segmentation}, various benchmark datasets such as \cite{yu2016modeling,nagaraja2016modeling} have been introduced to promote the development of the RES task. 
Aside from a few works \cite{hu2023beyond,liu2023gres} that consider real-world situations where one expression may refer to multiple or no targets, most prior research has revolved around the object-level notion that one expression refers to one target object. 
As shown in Table \ref{tab:dataset_compare}, the RefCOCO dataset \cite{yu2016modeling} is one of the most widely used grounding benchmarks to date, but it basically contains object-level visual and textual annotations. \textit{It does not take into account the part-level grounding task, which is necessary for future multimodal frameworks to perform as intelligent agents to realize the fine-grained perception of real world.} 
The existing grounding datasets do not support the training and evaluation of such advanced capabilities. 
Although some previous studies \cite{gong2017look,reddy2018carfusion,wah2011caltech,he2022partimagenet,mo2019partnet,chen2014detect,ramanathan2023paco} have made significant strides to develop high-quality part-level datasets, pushing towards fine-grained visual understanding.
However, these datasets focus on unimodal visual tasks and lack a deep connection between part-level masks and rich textual descriptions. 
\textit{To our knowledge, few studies have established this link. 
Consequently, there appears to be a limited availability of large-scale V-L datasets that facilitate part-level cross-modal understanding}, which is essential for two reasons. 
First, when describing objects, people always naturally gravitate towards detailing part-level properties, making part-level comprehension crucial for multimodal agents. Second, fine-grained part-level understanding can positively enhance object-level target perception, especially in challenging conditions like occlusion or deformation, which will consistently propel advancements in widely focused object-level grounding tasks. 
Thus, investigating ways to move beyond the current object-level constraints and advance towards finer-grained part-level grounding holds significant value and deserves thorough exploration.

\begin{figure*}[htbp]
    \centering
    \includegraphics[width=0.75\textwidth]{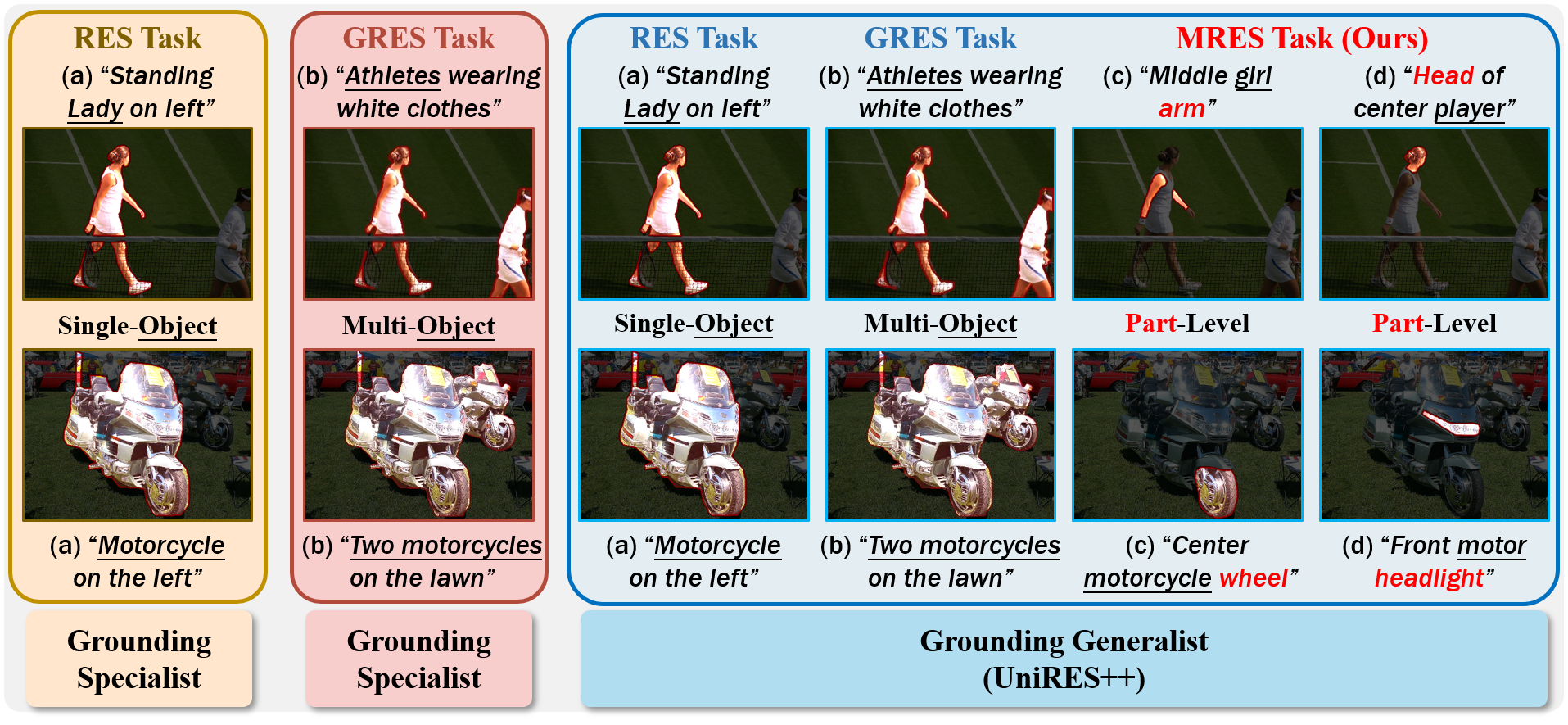}
    \vspace{-6pt}
    \caption{
    Classic {referring expression segmentation (RES)} and {generalized RES (GRES)} tasks only support expressions that indicate either a single target object (\eg, (a)) or multiple target objects (\eg, (b)), whereas our {multi-granularity RES (MRES)} task accommodates expressions that indicates the specific \textbf{\textit{part-level regions}} of object instances (\eg, (c)-(d) from our RefCOCOm benchmark). Besides, most previous grounding specialist models are designed exclusively for the object level grounding and they can only perform either RES or GRES. In contrast, with a unified architecture, our UniRES++ employs can simultaneously handle multiple localization tasks across various granularity levels.
    }
    \vspace{-6pt}
    \label{fig:intro}
\end{figure*}

2) \emph{Multi-Granularity Grounding Generalist Absence}: 
Since the inception of RES task proposed by \cite{hu2016segmentation}, various vision-language frameworks such as \cite{hu2016segmentation,wang2022cris,yang2022lavt,wang2024cm} have been devised to deal with the challenging feature extraction and alignment problems between visual and linguistic modalities, aiming to realize more precise RES. 
However, most previous methods have largely treated various levels of target granularity in isolation, primarily focusing on addressing object-level localization tasks. 
In other words, despite a few works \cite{hu2023beyond,liu2023gres} that consider real-world scenarios where a single expression may refer to multiple or even no targets, the majority of earlier approaches have been limited to designing specialist models tailored to specific RES tasks at the object level, lacking a generalist framework capable of simultaneously handling multi-granularity grounding tasks.
Such a framework is crucial for future multimodal models to act as intelligent agents capable of precise localization across different granularities in real-world deployments. 
Furthermore, there is intuitive synergy between multi-granularity RES tasks, where understanding at different levels can reinforce one another, but no prior work has explored the cooperative learning potential of multi-granularity data for multimodal models. 
\ul{In summary, while multi-grained RES is highly meaningful for multimodal embodied intelligence, there is a notable scarcity of research on visual grounding at multiple target granularity levels.
Besides, the current lack of relevant data and unified framework further compounds the challenge of this task}.

To this end, in this work, we attempt to fill the above mentioned important blank space, moving towards the unified multi-grained RES. 
This work is an extended version of our conference paper \cite{wang2024unveiling}, among which we attempt to promote the object-level RES task towards finer-grained vision-language understanding.
Specifically, to take a step further to finer-grained part-level RES, we put forward a new multi-granularity RES (MRES) task and a high-quality evaluation benchmark, RefCOCOm, by manually annotating part-level targets based on the existing RefCOCO benchmark with only object-level labels. Additionally, we construct by far the first grounding dataset to support part-level visual-textual annotations and also the largest-scale grounding dataset namely MRES-32M, comprising over 32.2M high-quality masks and captions on the provided 1M images.
In this extended version, we not only focus on understanding fine-grained part-level targets but also shift the research perspective towards the target granularity unified--multi-grained RES task. 
Since existing datasets and models mainly focus on designing grounding specialists for object-level target localization, lacking the necessary data resources and unified frameworks for the more practical multi-grained RES,
based on the finer-grained part-level grounding data collected from our conference version and the existing object-level localization data, we develop a new multimodal grounding generalist UniRES++ across omni-level target granularities. 
Since such a unified framework inherently requires rich local detail features to fulfill its multi-grained localization objectives, we start from a new perspective of structural designing, leveraging multi-granularity visual features to enrich the crucial detailed representations. 
As one of the first multimodal large language models (MLLMs) designed specifically centered on visual multi-granularity representations, UniRES++ incorporates targeted designs of Multi-Granularity Vision Flow for capturing multi-grained visual features and Multi-Granularity Feature Exploitation to achieve dynamic multi-grained feature interaction, enabling it to handle unified object-level and part-level RES tasks across omni-level visual granularities.
On existing benchmark datasets (\ie, gRefCOCO for {generalized RES (GRES)}, RefCOCO, RefCOCO+, RefCOCOg for classic RES task) and our RefCOCOm for MRES task, UniRES++ achieves new state-of-the-art (SOTA) performance with a shared architecture and parameters.

As shown in Fig. \ref{fig:intro}, the main contributions of this work can be summarized as follows:
\setlist{nolistsep}
\begin{itemize}[noitemsep,leftmargin=*]
    \item
    We construct a novel MRES task that bridges both object-level and part-level grounding, along with the corresponding evaluation benchmark RefCOCOm, pushing beyond classic object-focused RES task towards more comprehensive understanding of real-world scenes.
    \item
    To overcome the data scarcity in fine-grained V-L understanding, we build a multi-grained grounding dataset, named MRES-32M, which is the first grounding dataset to support part-level vision-language annotations and also the largest-scale grounding dataset to date.
    \item
    By incorporating our Multi-Granularity Vision Flow for capturing detailed visual features and the Multi-Granularity Feature Exploitation for dynamic feature interaction, we develop a new grounding generalist MLLM, UniRES++, as the first unified RES framework capable of handling multi-grained RES tasks across omni-level target granularities.
    \item
    UniRES++ achieves new SOTA performance across existing and our newly constructed benchmarks, including RefCOCOm for MRES, gRefCOCO for GRES, RefCOCO, RefCOCO+ and RefCOCOg for classic RES, demonstrating its superiority in unifying multi-granularity visual grounding.
\end{itemize}

\section{Related Work}
\label{relatedwork}

\noindent \textbf{Referring Expression Segmentation.}
Since \cite{hu2016segmentation} first introduce the basic concept of RES task, following studies, including \cite{yu2018mattnet,ye2019cross,hu2020bi,huang2020referring,wang2022cris,wang2024cm}, mainly adhere to a two-step pipeline where textual and visual features are encoded independently, and fused multimodal features are derived from these unimodal representations for mask prediction.
On this basis, the representation quality of the acquired multimodal features becomes the key to determine model performance.
Therefore, subsequent research \cite{ye2021referring,liu2021cross,ding2021vision,yang2022lavt,feng2022referring,ding2022vlt,10694805,lai2024lisa,zou2023segment,zou2023generalized} has continuously focused on optimizing the multimodal representations.
With the rapid development of large language models (LLMs), several recent studies \cite{lai2024lisa, xia2024gsva} have attempted to integrate LLMs into the RES task, offering language outputs and pixel-level mask prediction. 
Moreover, unlike previous works that primarily focus on the classic RES task where a single text description typically refers to only one target, a number of recent works \cite{hu2023beyond,liu2023gres} have focused on addressing the limitations of existing benchmark datasets \cite{kazemzadeh2014referitgame,yu2016modeling,nagaraja2016modeling,wu2020phrasecut}, paying attention to scenarios that are more reflective of real-world conditions where a single text description may refer to multiple or no corresponding target.
However, it's worth noting that most previous studies have focused on traditional object-level RES, with limited attention to the significant challenge of finer-grained part-level and unified multi-grained visual grounding, which remains underexplored in current research landscape.

\noindent \textbf{Multimodal Large Language Models.}\,
Among the rapidly advancing MLLMs recently, LLaVA \cite{liu2024visual,liu2024improved} stands out as one of the pioneering works to apply instruction tuning with vision information on LLMs like LLaMA \cite{touvron2023llama} and Vicuna \cite{chiang2023vicuna}.
By aligning vision with language modality, MLLMs can acquire notable image-level perception capabilities. 
On this basis, MLLMs can be extended to possess visual grounding ability by leveraging the templated supervision signal from bounding boxes \cite{chen2023shikra,chen2023minigpt}, achieving promising object-level perception. 
To realize pixel-level RES, LISA \cite{lai2024lisa} equips MLLMs with a pixel decoder, incorporating an additional [SEG] token to generate segmentation outputs. 
GSVA \cite{xia2024gsva} enhances LISA by introducing a [REJ] token to achieve finer-grained perception with robust segmentation results. 
However, most of existing RES research predominantly focuses on designing specialist models solely for object-level grounding task, lacking a generalist framework capable of simultaneously handling multi-granularity RES tasks.
This limitation stems from both the lack of multi-grained training data and tailored designs in model architectures for exploiting multi-grained visual representation.

\noindent \textbf{Fine-Grained Visual Perception.} 
The growing interest in fine-grained object understanding has led to the development of part-level annotated datasets across both specialized and general domains. 
Early works have pioneeringly introduced part-level datasets focused on specific categories, such as human body parts \cite{gong2017look}, animal parts \cite{reddy2018carfusion}, and vehicle components \cite{wah2011caltech}. 
In contrast, more general datasets like Pascal-Part \cite{chen2014detect}, PartNet \cite{mo2019partnet}, and PartImageNet \cite{he2022partimagenet} provide part annotations for various common objects. 
Recent works include a finer-grained panoptic segmentation dataset \cite{de2021part}, and PACO \cite{ramanathan2023paco}, which enhances object-level datasets by incorporating part and attribute annotations for finer-grained segmentation. 
Besides, VLPart \cite{sun2023going} further introduces a pipeline for part segmentation task and an effective detector capable of segmenting both open-vocabulary objects and their parts.
However, these datasets are primarily designed for visual perception tasks, lacking a strong connection between fine-grained part-level masks and detailed textual descriptions.

In the following, we will first introduce our constructed MRES task and evaluation benchmark RefCOCOm in Sec.~\ref{Multi-Granularity Visual Grounding Benchmark RefCOCOm}, and then present the collection pipeline and detail information for our MRES-32M data in Sec.~\ref{Multi-Granularity Visual Grounding Dataset MRES-32M}, which is followed by the elaboration about our well designed grounding generalist UniRES++ in Sec.~\ref{UniRES++_Model} and the experimental results in Sec. \ref{experimental_results}.

\section{Multi-Granularity Grounding Benchmark}
\label{Multi-Granularity Visual Grounding Benchmark RefCOCOm}

\subsection{Multi-Granularity RES Task}

The interaction capability of a multimodal agent hinges on the visual granularity it can perceive and comprehend. 
However, in the current research landscape, multi-granularity visual grounding remains neglected and under-explored. 
To overcome the limitation where visual-language alignment is confined to the object level, we aim to advance towards finer-grained vision-language understanding. 
Specifically, we introduce the multi-granularity RES task, requiring models to uniformly ground entities across both part and object levels in response to diverse textual references.
Within a single modality, this calls for an improved ability to grasp complex, implicit knowledge structures, \eg, the granularity outlined by textual descriptions and the heightened precision of pixel-level semantic understanding.
Across the vision-language modalities, finer-grained references present new challenges for the accuracy and robustness of cross-modal alignment.
Furthermore, as prior studies are mainly restricted to object-level referring segmentation and no benchmarks exist for evaluating part-level grounding, we establish a new benchmark named RefCOCOm, which will be detailed below.

\subsection{RefCOCOm Benchmark}

\noindent \textbf{Manual Annotation Pipeline.}
To facilitate seamless evaluation for community researchers without code modifications and to build a high-quality multi-granularity benchmark, we manually conduct part-level annotations on three official subsets (\ie, validation, testA, and testB) of the most widely used RefCOCO benchmark dataset \cite{yu2016modeling} by splitting existing object masks.
We first carefully establish the criteria for delineating part masks and identify the necessary part categories for each object. 
Then, SAM \cite{kirillov2023segment} is used to obtain initial segmentation results, which are refined and captioned by 30 skilled annotators via an online annotation tool developed by ourselves.
To ensure annotation quality, we conduct spot checks at three progress nodes (\ie, 20\%, 50\%, and 80\% of the total workload), providing corrective feedback as needed.

\begin{figure}[thbp]
    \centering
    \includegraphics[width=0.48\textwidth]{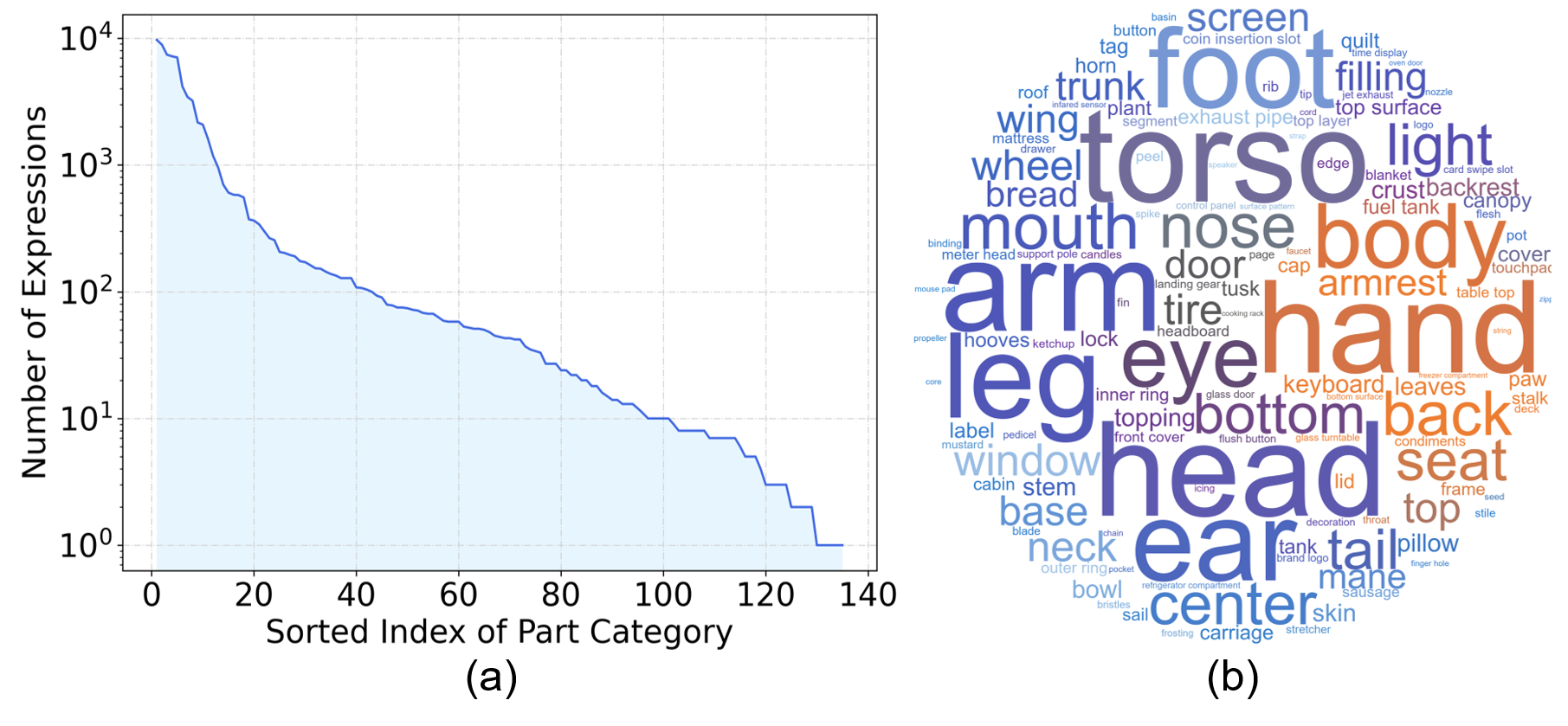}
    \vspace{-10pt}
    \caption{RefCOCOm benchmark statistics.
    (a) the number of referring expressions per parts' category in the log scale. 
    (b) the word cloud highlights the head categories.
    }
    \label{fig_RefCOCOm}
    \vspace{-5pt}
\end{figure}

\begin{figure}[htbp]
    \centering
    \includegraphics[width=0.48\textwidth]{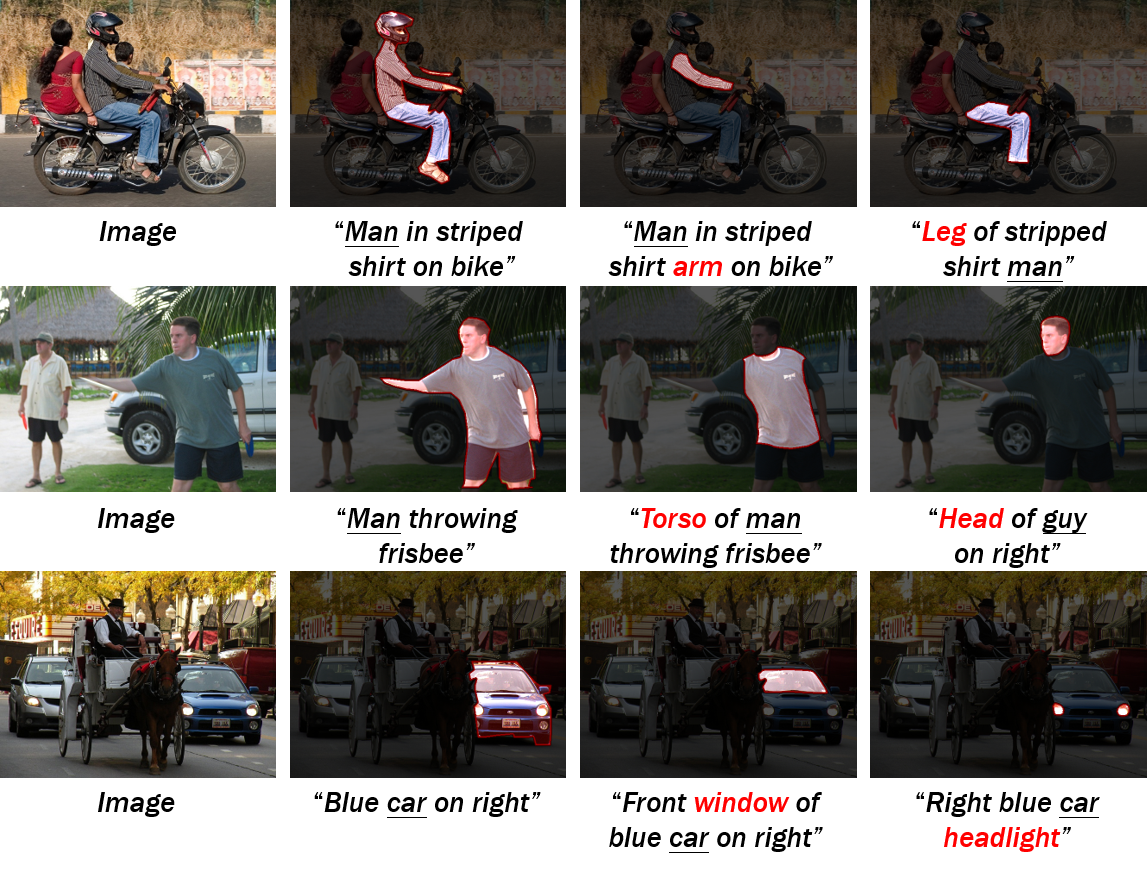}
    \vspace{-10pt}
    \caption{
    Selected samples from our proposed RefCOCOm benchmark for {multi-granularity RES (MRES)} task. 
    }
    \label{fig_examplesofRefCOCOm}
    \vspace{-10pt}
\end{figure}

\noindent \textbf{Benchmark Details.}
In total, we collect 70k part references with 26k corresponding masks. 
The referring expressions in our RefCOCOm benchmark have an average length of 5.1 words, spanning 80 object categories and 391 part categories.
The number of referring expressions per part category and a word cloud highlighting the major categories are shown in Fig. \ref{fig_RefCOCOm}.
Combined with the original object-level annotations, RefCOCOm includes 34k masks and 92k references overall. 
As an extended version of RefCOCO \cite{yu2016modeling} dataset, RefCOCOm sets higher demands for referential comprehension and fine-grained visual perception abilities.
Additionally, we provide more examples in our RefCOCOm benchmark for the proposed MRES task in Fig. \ref{fig_examplesofRefCOCOm}. 
To align with the goal of multi-granularity unification, we use mean Intersection-over-Union (mIoU) as the evaluation metric.

\begin{figure*}[htbp]
    \centering
    \includegraphics[width=0.75\textwidth]{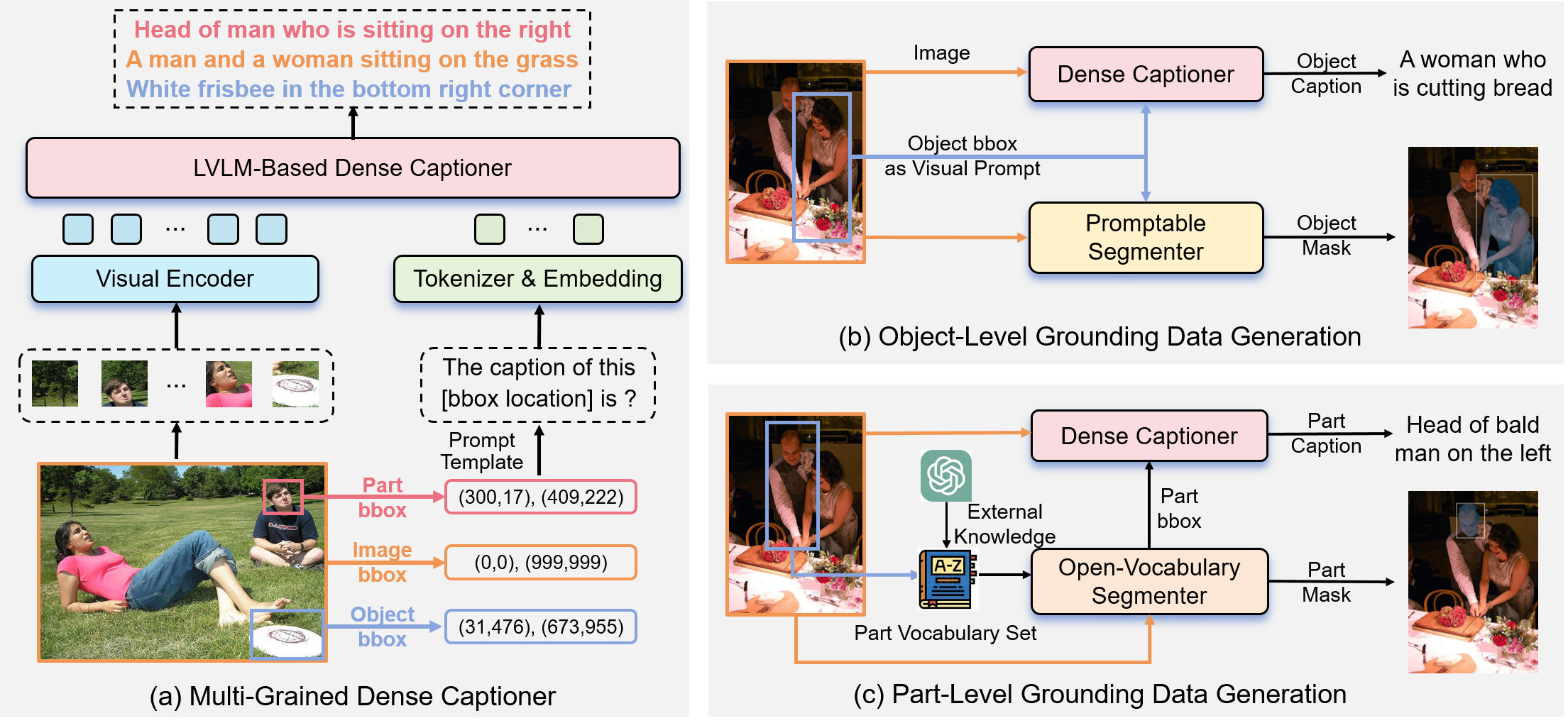}
    \vspace{-10pt}
    \caption{
    The illustration of our data engine for constructing the MRES-32M dataset.
    (a) We begin by fine-tuning an MLLM to create a robust dense captioner capable of handling captioning across three granularity levels.
    (b) For object-level grounding data, we feed images and bounding boxes into the captioner and a powerful segmenter, generating captions and masks for various objects.
    (c) Leveraging LLMs' external knowledge, we decompose existing object category annotations into a vocabulary of part-level tags, which are then processed by an open-vocabulary segmenter and our captioner to obtain part-level annotations.
    }
    \label{fig_dataengine}
    \vspace{-10pt}
\end{figure*}

\section{Multi-Granularity Grounding Dataset}
\label{Multi-Granularity Visual Grounding Dataset MRES-32M}

\subsection{Data Collection Engine}

Due to the inherent complexity of classic RES task, the associated training data demands significant annotation costs across both textual and visual domains. 
When extended to part-level granularity, these challenges become even more exacerbated.
We argue that the main barrier to open-world grounding is the limited scalability of existing data. 
By harnessing powerful foundation models for synergistic improvement, we present an advanced data engine that can automatically generate reliable visual grounding data.

\noindent \textbf{Multi-Grained Dense Captioner Acquiring.} 
As large language models \cite{chiang2023vicuna,touvron2023llama} continue to enhance the multimodal domain with profound capabilities, MLLMs \cite{bai2023qwen,chen2023minigpt,chen2023shikra} have effectively expanded into the grounding domain. These models use numerical coordinates to represent bounding boxes, integrating positional information into the language model via word embeddings for contextual understanding. However, the corresponding exploration into part granularity remains nascent. Constrained training data have restricted MLLMs' fine-grained descriptive capabilities solely to the object level. To fully leverage the open-world visual knowledge acquired from extensive pretraining on image-level and object-level data, we devise a unified fine-tuning scheme that enables MLLMs to function as generalist captioners across all granularity levels, as shown in Fig. \ref{fig_dataengine} (a). 

All data for these granularity levels are sourced from manual annotations for reliability. The input includes an image and the corresponding bounding box, with coordinates normalized to integers within the [0,999] range.
For image granularity, we use the COCO dataset \cite{lin2014microsoft}, where all bounding boxes are uniformly represented as $(0,0),(999,999)$. 
For object granularity, we rely on the Visual Genome dataset \cite{krishna2017visual}. 
For part granularity, we draw on unimodal semantic segmentation data \cite{ramanathan2023paco,he2022partimagenet,chen2014detect} and construct dense captions using the template $PartNameX\; of\; ObjectNameY$.
This unified multitask training approach fosters synergy across different granularity levels, allowing MLLMs to incorporate more comprehensive and detailed information to enrich part-level descriptions. At the same time, part granularity knowledge aids in refining object-level understanding.

\noindent \textbf{Model-Assisted Data Generation.} 
With the obtained Multi-Grained Dense Captioner mentioned above, we intend to collect multi-grained grounding data across both object and part granularity levels.
For object-level grounding data generation, we utilize the large-scale object detection dataset Object365 \cite{shao2019objects365} to provide highly reliable bounding boxes. Its rich category labels ensure comprehensive knowledge coverage. As depicted in Fig. \ref{fig_dataengine} (b), these bounding boxes serve as visual prompts, which are independently processed by both a promptable segmenter (\ie, segment anything model \cite{kirillov2023segment}) and our generalist dense captioner to obtain segmentation masks and detailed semantic descriptions, separately.

For part-level grounding data, we introduce a hierarchical annotation scheme built upon existing object-level annotations, as presented in Fig. \ref{fig_dataengine} (c). 
Specifically, we leverage GPT-4 \cite{openai2023gpt} with extensive external knowledge to decompose objects in the image and generate a customized part vocabulary set. This tailored vocabulary set is then input into an open-vocabulary segmenter \cite{sun2023going}, producing precise part masks and bounding boxes. These bounding boxes are subsequently fed into our dense captioner for obtaining detailed captions.

\noindent \textbf{Data Filtering.} 
To ensure the high quality of our MRES-32M's data as much as possible, after completing the multi-granularity annotation for all the images, we utilize CLIP model \cite{clip} for effective data filtering. The bounding box is cropped from the original image and sent to the clip encoder along with the dense caption to measure V-L semantic similarity. 
To maintain consistency between visual and linguistic annotations, we retain box-caption pairs with a similarity score above 0.5 as the final annotations.

\subsection{MRES-32M Dataset Details}

\vspace{-6pt}
\begin{table}[htbp]
    \centering
    \small
    \setlength{\tabcolsep}{2.25pt} 
    \caption{Comparisons with previous object-level visual grounding datasets and part-level segmentation datasets. \# denotes the specific number, 
    where Cats and Avg Len denote the object/part categories and the average length of referring expressions.
    ``-'' denotes the corresponding part masks or captions are unavailable. }
    \begin{tabular}{lccccc}
    \specialrule{.1em}{.05em}{.05em}
        Dataset & \#Imgs & \#Objs & \#Parts & \#Cats & \#Avg Len \\
        \midrule
        \multicolumn{6}{l}{\color{gray} Object-Level Visual Grouding} \\
        ReferIt \cite{kazemzadeh2014referitgame} & 20K & 97K & -- & 238/-- & 3.2 \\
        RefCOCO \cite{yu2016modeling} & 20K & 50K & -- & 80/-- & 3.6 \\
        RefCOCO+ \cite{yu2016modeling} & 20K & 49K & -- & 80/-- & 3.5 \\
        RefCOCOg \cite{nagaraja2016modeling} & 26K & 54K & -- & 80/-- & 8.4 \\
        gRefCOCO \cite{liu2023gres}& 20K & 60K & -- & 80/-- & 3.7 \\
        \midrule
        \multicolumn{6}{l}{\color{gray} Part-Level Segmentation \& Detection} \\
        PartsIN \cite{he2022partimagenet}  & 24K & 24K & 112K & 158/609 & -- \\
        PascalPart \cite{chen2014detect} & 19K & 40K & 363K & 20/193 & -- \\
        PACO \cite{ramanathan2023paco} & 20K & 260K & 641K & 75/456 & -- \\
        \midrule
        \multicolumn{6}{l}{\color{gray} Multi-Grained Visual Grounding} \\
        \rowcolor{mygray}
        MRES-32M (Ours) & \textbf{1M} & \textbf{15.3M} & \textbf{16.9M} & \textbf{365/2299} & 4.6 \\
    \specialrule{.1em}{.05em}{.05em}
    \end{tabular}
    \vspace{-5pt}
    \label{tab:dataset}
\end{table}

\begin{figure*}[htbp]
    \centering
    \includegraphics[width=0.80\textwidth]{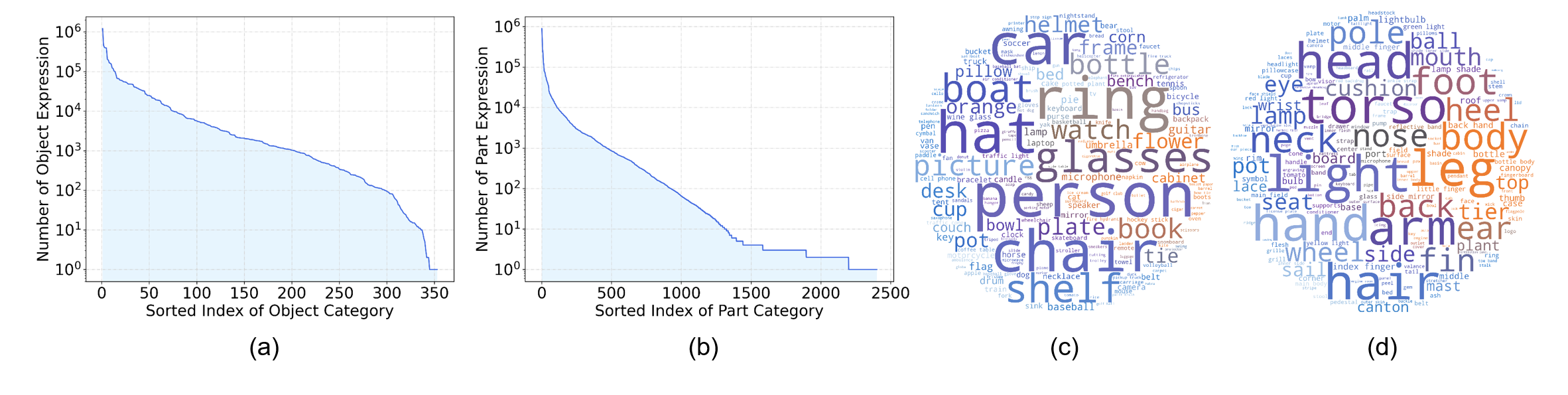}
    \vspace{-15pt}
    \caption{
    MRES-32M dataset statistics.
    (a) the number of referring expressions per objects' category in the log scale. 
    (b) the number of referring expressions per parts' category in the log scale. 
    (c) the word cloud highlights the head objects' categories. 
    (d) the word cloud highlights the head parts' categories. 
    }
    \label{fig_analysisofMRES-32M}
    \vspace{-3mm}
\end{figure*}

\begin{figure}[htbp]
    \centering
    \includegraphics[width=0.48\textwidth]{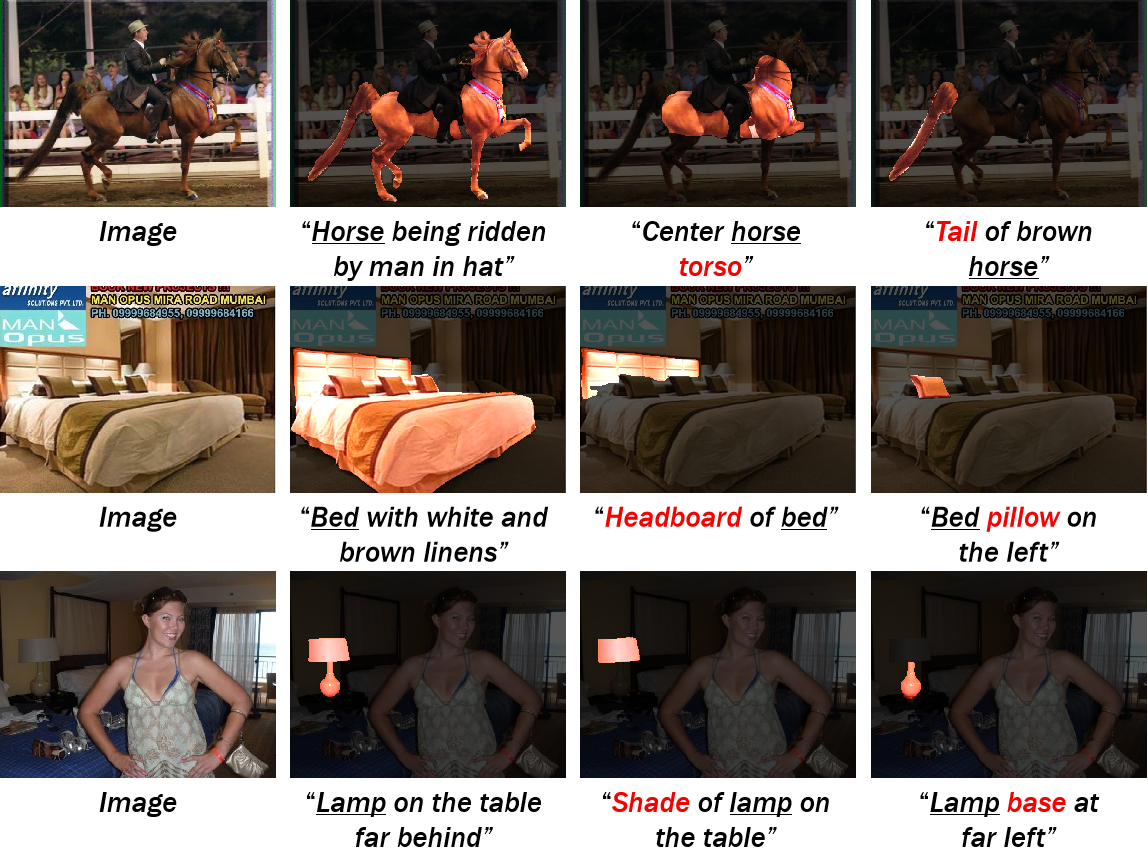}
    \vspace{-10pt}
    \caption{Selected samples from our well constructed MRES-32M dataset.}
    \label{fig_examplesofMRES-32M}
\end{figure}

As shown in Table \ref{tab:dataset}, existing benchmark datasets, such as the most widely used RefCOCO \cite{yu2016modeling}, are limited by their small scale and lack of part-level dense annotations. PACO \cite{ramanathan2023paco}, a pioneering work in finer-granularity unimodal visual comprehension, includes only semantic tags for objects or parts, without detailed descriptions or visual context as referring expressions.
In summary, we compare the proposed MRES-32M with existing datasets and highlight its unique and significant features in Table \ref{tab:dataset}. 
Our MRES-32M dataset consists of 365 object categories and 2,299 associated part categories. Compared to existing datasets, it covers a broader range of multimodal knowledge, marking a significant step towards open-world understanding. In Fig. \ref{fig_analysisofMRES-32M}, we present the number of referring expressions per object and part category, along with a word cloud highlighting the key object and part categories, respectively.
Additionally, a few examples from our MRES-32M dataset are provided in Fig. \ref{fig_examplesofMRES-32M}.

\noindent \textbf{Unified Multi-Granularity.} Compared to the previous grounding counterparts, our MRES-32M is the first visual grounding dataset to cover both part and object granularity. In comparison with part-level visual segmentation datasets, MRES-32M offers informative and unique fine-grained descriptions for each part mask. \\
\noindent \textbf{More Complex References.} Benefiting from our MLLM-based generalist dense captioner, the references in our MRES-32M are more fully integrated with visual context for entity (\ie, part and object) descriptions. Without adhering to a fixed template, entities' relationships and attributes are highlighted in free-form natural language expressions. \\
\noindent \textbf{More Diversified Categories.} MRES-32M consists of 365 object categories and 2,299 associated part categories. In contrast with existing grounding datasets, it covers a broader range of multimodal knowledge, making it a significant step toward open-world understanding. \\
\noindent \textbf{Breakable Data Scales.} MRES-32M is by far the largest dataset in the current visual grounding research community, surpassing the largest existing grounding dataset, RefCOCOg \cite{nagaraja2016modeling}, by 38 times in terms of image count and 283 times in object instance count. Besides, our MRES-32M also contains 58 times more part instances than the largest part semantic segmentation benchmark dataset PACO \cite{ramanathan2023paco}.

\definecolor{lightblue}{RGB}{30,144,255}
\begin{figure*}[htbp]
    \centering
    \includegraphics[width=0.82\textwidth]{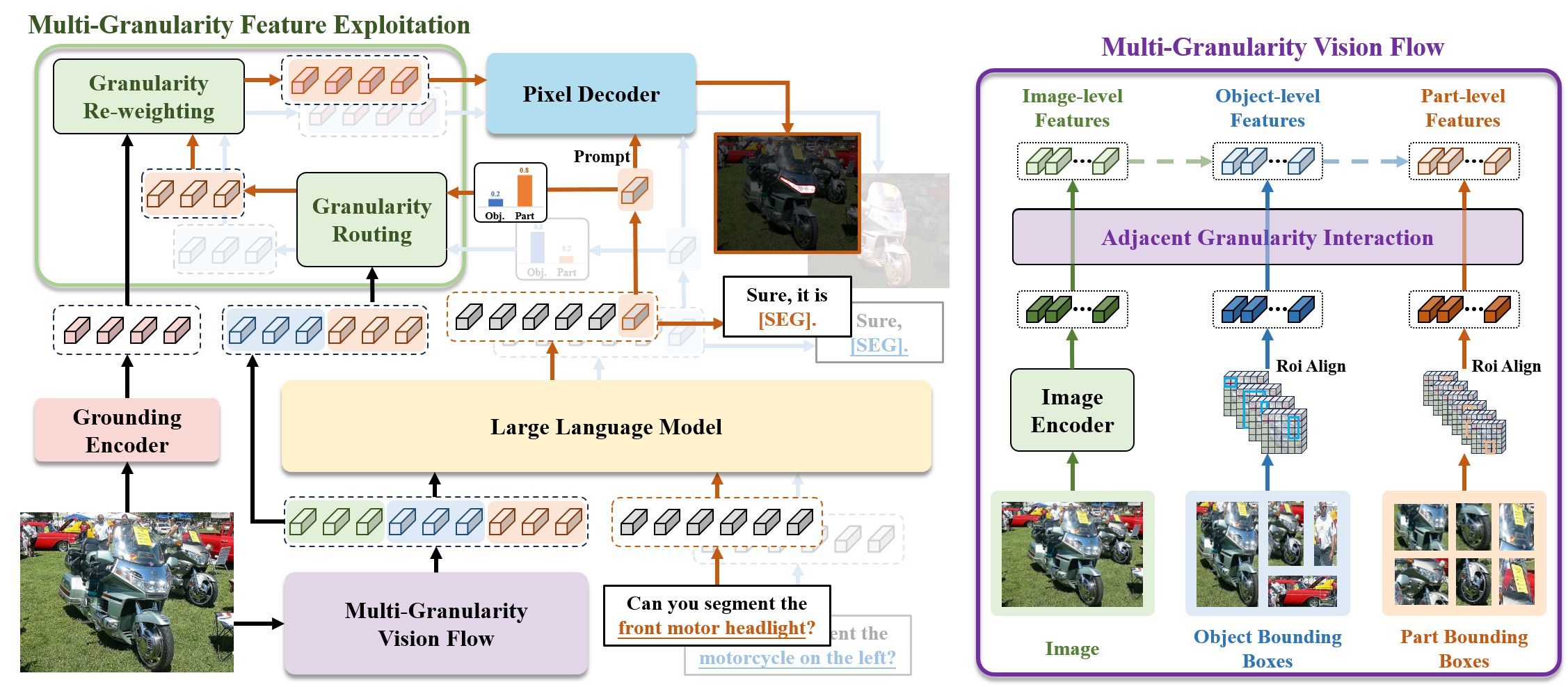}
    \vspace{-5pt}
    \caption{
    The architecture of our UniRES++ as the unified framework for multi-grained RES tasks across various target granularities.
    The foreground entity and the blurred background respectively represent the processes of UniRES++ in achieving part-level (as highlighted in \textbf{\textcolor{orange}{orange}}) and object-level (as highlighted in \textbf{\textcolor{lightblue}{lightblue}}) target localization.
    As one of the first MLLMs designed specifically with multi-grained visual features for fine-grained V-L understanding, UniRES++ mainly comprises several parts: Multi-Granularity Vision Flow for capturing multi-grained visual features, Grounding Encoder to obtain foundational visual feature representations, LLM, Multi-Granularity Feature Exploitation for dynamic feature selection, and Pixel Decoder to generate the segmentation mask.
    }
    \label{fig_model}
    \vspace{-10pt}
\end{figure*}

\section{Multi-Granularity Grounding Model}
\label{UniRES++_Model}

After detailing our efforts to push the traditional RES paradigm towards finer-grained V-L understanding through attempts in task formulation, evaluation benchmark and large-scale training data, we now introduce the multi-granularity RES model UniRES++, designed to advance unified RES across hierarchical visual granularities.
Our primary goal is to develop a unified grounding framework capable of locating objects at multiple levels of granularity. 
Such a framework inherently requires rich local detail features to fulfill its localization objectives. 
Currently, the mainstream research on MLLM's architecture focuses on encoding multiple low-resolution image patches split from high-resolution source images to capture detailed information \cite{li2024monkey,lin2023sphinx}. 
However, this paradigm introduces unignorable computational overhead during inference, which appears to be suboptimal. 

Therefore, in this work, we start from a new perspective, incorporating multi-granularity visual features with powerful LLMs to enrich the crucial detailed representations. 
Considering that the visual feature representations required by multimodal agents to complete multi-grained localization tasks vary in practical scenarios, we introduce the tailored design for dynamic interaction of multi-granularity features after effectively capturing them. 
This allows for the dynamic and appropriate utilization of these fine-grained features.
Benefiting from our introduced multi-granularity visual feature acquisition and dynamic interaction, UniRES++ can autonomously adjust the dominance of each granularity of the captured features based on the target granularity, thereby achieving accurate multi-granularity target localization.
As one of the first MLLMs designed specifically with multi-granularity visual representations, our UniRES++ architecture is depicted in Fig. \ref{fig_model}. 
The architecture comprises several key components: Multi-Granularity Vision Flow (MGVF) for capturing multi-grained visual features, Grounding Encoder, LLM, Multi-Granularity Feature Exploitation (MGFE), and Pixel Decoder to generate the final mask outputs. 
In the following, we will first provide an overview of UniRES++ structure, then delve into the key components (\ie, MGVF and MGFE). Finally, we will discuss the training and inference strategies employed for UniRES++.

\subsection{Overall Architecture}
\label{sec:method1}

The overall architecture of UniRES++ is depicted in Figure \ref{fig_model}. 
It is designed to seamlessly integrate image and text inputs for advanced cross-granularity feature extraction and target localization. 
Specifically, it begins with the Grounding Encoder (\ie, a pre-trained SAM encoder \cite{kirillov2023segment}), which compresses the input image into foundational visual features with rich representations. Simultaneously, original image is fed into the MGVF to extract diverse multi-granularity visual features, including three distinct levels: image, object and part. 
The constructed vision flow initially captures these multi-grained features and then dynamically employs adjacent-level feature interactions, from coarse image-level to fine-grained part-level, to enhance the representations across all levels of granularity. 
Through unidirectional guided transitions, the model facilitates dynamic interplay among the different levels of fine-grained visual features captured within the image. 
This tailored design will be further explained in Sec. \ref{sec:method2}.

Subsequently, the obtained multi-granular visual tokens are first aligned through a projection layer and then input into the powerful LLM, accompanied by language descriptions that pinpoint the targets at various granularities. 
Through effective multimodal sequence modeling, the LLM outputs a special token, [SEG]. 
Guided by our targeted training, the [SEG] token will convey the implicit information about the target granularity that the LLM autoregressively determines is required for the current task. 
Following this, UniRES++ selects the feature representations of the specified granularity from multi-grained sequence, which has been obtained through unidirectional guidance, based on the granularity determined by the current [SEG].
The dominant granularity features once identified are used to specifically enhance the low-level details within the visual representations encoded by Grounding Encoder, refining the foundational visual features and emphasizing details pertinent to the target region.
The [SEG] token, together with the enhanced visual features, is then conveyed to the Pixel Decoder. The decoder utilizes these enriched representations to generate a precise mask prediction for the visual target, effectively aligning with the granularity required by the [SEG] token and ensuring the accurate localization.
Detailed explanations are provided in Sec. \ref{sec:method3}.
In summary, UniRES++ not only ensures precise feature extraction and interaction but also facilitates adaptive responses to varying target granularity requirements, making it highly effective for detailed and context-sensitive grounding tasks.

\subsection{Multi-Granularity Vision Flow}
\label{sec:method2}

To realize multi-granularity grounding tasks within a single unified MLLM framework, it is essential to consider both the acquisition of multi-grained features and their subsequent adaptive exploitation. 
The MGVF is specifically designed as a key component for pursuing the unified RES, effectively obtaining the multi-grained visual representations. 
Unlike directly feeding high-resolution images into the model to capture detailed information, our MGVF part uses the encoded features of low-resolution images as the base visual features. It then effectively supplements fine-grained visual features by incorporating features from the selected regions of interest in the encoded features of high-resolution images, which contain richer details. This enables the efficient and comprehensive acquisition of fine-grained visual features.

Specifically, our MGVF starts with a image encoder $E_l$ (\ie, CLIP ViT-L/14 \cite{clip}) encoding input images $I_l$ at the low resolution (\ie, 336$\times$336) to capture basic image-level features $F_l$:
\begin{equation} 
    F_l = E_l (I_l) \in \mathbb{R}^{N_l \times C}
\end{equation} 
where $N_l$ is the number of patches specific to a fixed low resolution and $C$ is the feature dimension. Moreover, UniRES++ aims to efficiently incorporate additional detailed features from object and part regions as auxiliary information. Initially, images $I_l$ resized to high-resolution ones $I_h$, containing more detailed information, are encoded by the CLIP ConvNeXt-L \cite{yu2024inceptionnext}, a visual encoder $E_h$ under the convolution paradigm that inherently supports variable resolution inputs:
\begin{equation} 
    F_h = E_h(I_h) \in \mathbb{R}^{N_h \times C}
\end{equation}
where $N_h$ is the number of patches specific to a fixed high resolution. Following this, the introduced open-domain instance detectors \cite{minderer2024scaling,sun2023going} take the original images $I_l$ as input to generate instance proposals at the object-level and part-level. Let $\mathcal{S}_o =  \{ o_i \mid i = 1, 2, \dots, N_o \}$ and $\mathcal{S}_p = \{ p_i \mid i = 1, 2, \dots, N_p \}$ denote the sets of proposal bounding boxes for objects and parts, respectively.
These acquired proposal bounding boxes are then used to perform ROI Align \cite{he2017mask} on the encoded features of the high-resolution images $I_h$, thus extracting the object-level and part-level features and maintaining the total length of the visual token sequence within a manageable range. Specifically, for each object bounding box in $\mathcal{S}_o$, the object-level feature $F_{o_i}$ is extracted as: 
\begin{equation} 
    F_{o}^i = \text{ROI Align}(F_h, o_i) \in \mathbb{R}^{C}
\end{equation} 
and similar operation is adopted for each part bounding box:
\begin{equation} 
    F_{p}^i = \text{ROI Align}(F_h, p_i) \in \mathbb{R}^{C}
\end{equation}
Thus, we obtain sets of object-level features $F_o = \left[ F_{o}^1, F_{o}^2, \dots, F_{o}^{N_o} \right] ^ \top \in \mathbb{R}^{N_o \times C}$ and part-level features $F_p = \left[ F_{p}^1, F_{p}^2, \dots, F_{p}^{N_p} \right] ^ \top \in \mathbb{R}^{N_p \times C}$ . This paradigm not only enhances detail representation but also optimizes computational resource usage by selectively processing areas of interest. 

The derived multi-grained visual features (\ie, image-level $F_l$, object-level $F_o$, and part-level $F_p$) are then seamlessly integrated into the next phase of dynamic feature interaction. 
Specifically, we utilize the effective cross-attention mechanism to first enhance the finer-grained object-level feature with the image-level feature, where $F_o$ serves as the query and $F_l$ is used as both the key and value: 
\begin{equation} 
    \tilde{F}_o = \text{CrossAttn}(F_o, F_l, F_l) \in \mathbb{R}^{N_o \times C}
\end{equation}
please note that in the subsequent cross-attention operation, the same order of (query, key, value) is applied. Then the enhanced object-level detail representation is further utilized to strengthen the further finer-grained part-level region features:
\begin{equation} 
    \tilde{F}_p = \text{CrossAttn}(F_p, \tilde{F}_o, \tilde{F}_o) \in \mathbb{R}^{N_p \times C}
\end{equation}
thereby obtaining the final output of three granularity features by the MGVF. 
In this manner, the adjacent-level interaction scheme, realized through unidirectional guided transitions from coarse image-level to fine-grained part-level, facilitates dynamic interplay among the different levels of fine-grained visual features captured within the image.

\subsection{Multi-Granularity Feature Exploitation}
\label{sec:method3}

We design the UniRES++ architecture based on the principle of first capturing rich multi-granularity features, then dynamically adjusting the internal utilization of these features according to the specific granularity required for multi-grained target localization.
Considering that localization tasks at different target granularities require varying levels of fine-grained features, it is necessary for our unified UniRES++ to introduce dynamic adjustment of the tailored utilization of fine-grained features at different levels.
The above section has detailed how we leverage the designed Multi-Granularity Vision Flow to obtain feature representations rich in both global and local detail information. This section will further elucidate how, after obtaining multi-granularity features guided by adjacent levels, we dynamically utilize them according to the specific target granularity of the localization task.

\textbf{Granularity Decoupling of [SEG] Token}: Specifically, the obtained multi-granular visual tokens first pass through a vision-text projection layer $P$ and are then fed into the powerful LLM, along with the language descriptions that indicate the corresponding targets at various granularities. The projected visual tokens can be denoted as: 
\begin{equation}
    F_v = P([F_l || \tilde{F}_o || \tilde{F}_p]) \in \mathbb{R}^{(N_l + N_o + N_p) \times D}
\end{equation}
where $\|$ represents the concatenation operation and $D$ is the unified feature dimension after projection. Through effective multimodal sequence modeling, the LLM outputs a special token, namely [SEG]. Unlike the previous model LISA \cite{lai2024lisa}, 
which introduces a single special [SEG] token indicating the multimodal features autoregressively generated by the LLM, 
we extend [SEG\_OBJECT] and [SEG\_PART] into the vocabulary of LLM's tokenizer in UniRES++, forcibly pursuing the granularity decoupling of [SEG] token $C_g$: \begin{equation}
    \hat{g} = C_g(\text{[SEG]}) \in {0,1} 
\end{equation} 
where $\hat{g} = 0$ indicates it corresponds to [SEG\_OBJECT] token which represents  object-level (\ie, multi-object \& single-object) localization task and $\hat{g} = 1$ refer to [SEG\_PART] token, indicating a part-level localization task.  In this way, during the training phase, when UniRES++ performs object-level and part-level grounding tasks, explicit supervision of the [SEG] token category enables the model to establish a deep connection between the current text description and the given label mask, thus forcing the model to become self-aware of the target granularity of the current localization task, \ie, task granularity awareness. The loss function of the text generation is directly incorporated to supervise this special [SEG] token.

\textbf{Granularity Routing \& Re-Weighting}: Based on the routing guidance provided by [SEG], our UniRES++, upon detecting the required granularity of the target to be localized, selects either the mutually enhanced object-level or part-level features for further processing. Specifically, we define the routing process of obtaining the selected features $\tilde{F}$ as:
\begin{equation} \tilde{F} = (1 - \hat{g}) \cdot \tilde{F}_o + \hat{g} \cdot \tilde{F}_p
\end{equation} 
where $\hat{g}$ acts as a selector between the object-level features $\tilde{F}_o$ (\ie, $\hat{g}=0$) and the part-level features $\tilde{F}_p$ (\ie, $\hat{g}=1$).
While the feature representations $F_g$ encoded by the Grounding Encoder are semantically rich, they lack targeted detail enhancement. Therefore, the selected features at a specific level are utilized to re-weight the encoded visual representations from the Grounding Encoder, refining the foundational visual features by emphasizing details relevant to the target region. Notably, this re-weighting method effectively employs cross-attention, which can be expressed as follows:
\begin{equation} F_{r} = \text{CrossAttn}(F_g, \tilde{F}, \tilde{F}) \in \mathbb{R}^{N_g \times C}
\end{equation} 
where $F_r$ are the re-weighted visual features, $N_g$ are the number of patches of image encoded by the Grounding Encoder.
Then, the [SEG] token, carrying multimodal representations about the target granularity, along with the adjacent-level guided visual features at the corresponding granularity level, is passed to the Pixel Decoder. The decoder utilizes these comprehensive representations to generate a precise mask prediction of the visual target, effectively matching the required granularity as specified by the [SEG] token.
In this manner, the entire framework dynamically adjusts the dominance of the granularity-specific features within the multi-granularity feature set based on the LLM's determination of the current granularity needed for target localization.

\subsection{Training \& Inference Strategy}
\label{sec:method4}

\noindent \textbf{Training.} Building upon the unified framework UniRES++ for multi-granularity grounding tasks, a well-designed training strategy is crucial to achieve the expected effects. 
The entire training process of UniRES++ is structured into two phases. 
In the first stage, we incorporate the high-quality dataset GranD \cite{rasheed2024glamm} and follow the same training settings to pre-train our end-to-end grounding framework. 
It is noteworthy that decoupling the [SEG] tokens for object/part granularity level is not introduced during the pre-training phase.
Through this initial phase of learning from the knowledge-rich GranD data, our UniRES++ accumulates a vast repository of visual-language understanding, which lays a solid foundation for learning multi-granularity grounding tasks in the second phase.
Following this, the second training stage involves joint learning on multi-granularity grounding data. 
We train our grounding generalist using object-level data from existing datasets (\ie, RefCOCO \cite{yu2016modeling}, RefCOCO+ \cite{yu2016modeling}, RefCOCOg \cite{nagaraja2016modeling} and gRefCOCO \cite{liu2023gres}) and our multi-granularity MRES-32M dataset. 
Given that region-level captioning and fine-grained grounding tasks are essentially mutually reinforcing, region-level captioning datasets (\ie, VG \cite{krishna2017visual}, RefCOCO \cite{yu2016modeling}, RefCOCO+ \cite{yu2016modeling} and RefCOCOg \cite{nagaraja2016modeling}) are also integrated as part of the multi-task training data. 
During this training stage, the losses for multi-granularity grounding tasks and region captioning tasks are employed to optimize for each task's specific requirements, which can be expressed as a text generation loss \(\mathcal{L}_{\text{txt}}\) and a segmentation mask loss \(\mathcal{L}_{\text{mask}}\). The overall objective is a weighted sum of these losses:
\begin{equation}
    \mathcal{L} = \lambda_{lm} \mathcal{L}_{lm} + \lambda_{\text{mask}} \mathcal{L}_{\text{mask}}
\end{equation}
where \(\lambda_{lm}\) and \(\lambda_{\text{mask}}\) balance the contributions of the generation and segmentation losses, respectively. Specifically, \(\mathcal{L}_{lm}\) is the autoregressive cross-entropy loss for text generation:
\begin{equation}
    \mathcal{L}_{lm} = -\sum_{t=1}^T \log P(x_t \mid x_{<t})
\end{equation}
where $x_t$ represents the token at position $t$ within the sequence fed into the LLM.
For \(\mathcal{L}_{\text{mask}}\), it encourages high-quality segmentation results. To compute the segmentation losses, we combine per-pixel binary cross-entropy (BCE) loss and DICE loss, weighted by \(\lambda_{\text{bce}}\) and \(\lambda_{\text{dice}}\):
\begin{equation}
    \mathcal{L}_{\text{mask}} = \lambda_{\text{bce}} \, \text{BCE}(\hat{\mathbf{M}}, \mathbf{M}) + \lambda_{\text{dice}} \, \text{DICE}(\hat{\mathbf{M}}, \mathbf{M})
\end{equation}
\begin{equation}
    \small
    \text{BCE} = -\frac{1}{N} \sum_{i=1}^{N} \left[ M_i \log (\hat{M}_i) + (1 - M_i) \log (1 - \hat{M}_i) \right]
\end{equation}
\begin{equation}
    \small
    \text{DICE} = 1 - \frac{2 \sum_{i=1}^{N} \hat{M}_i M_i}{\sum_{i=1}^{N} \hat{M}_i + \sum_{i=1}^{N} M_i}
\end{equation}
where \(\hat{\mathbf{M}}\) are the model's predictions and \(\mathbf{M}\) are the ground-truth masks, respectively. The total number of pixels is $N$.

\noindent \textbf{Inference.} Furthermore, UniRES++ exhibits some variations in the inference phases for multi-granularity grounding tasks and region-level captioning task. 
When executing the multi-granularity target localization tasks, which are the primary focus of this paper, UniRES++ takes as input the text descriptions that refer to targets of specific granularities along with the original image. 
It will first acquire the multi-granular visual features and determine the target granularity level using the LLM. 
Then, it emphasizes the exploitation of the corresponding granularity features within the multi-granularity vision flow, as judged by the LLM. 
Finally, under the guidance of the [SEG] token specific to the task granularity, the pixel decoder generates the final target segmentation mask. 
When UniRES++ performs V-L understanding tasks other than grounding, such as region captioning task, the inference process slightly differs. 
Specifically, an additional visual encoder is introduced to encode the region areas, while other aspects, such as the acquisition of multi-granular visual features, remain similar to those in grounding tasks. Based on this, all visual features, along with textual instructions, are fed into the strong LLM, which will output the expected dense caption for the original image in an autoregressive manner.

\section{Experimental Results}
\label{experimental_results}

To evaluate the effectiveness of our UniRES++ framework, comprehensive experiments are conducted on three classic RES datasets (\ie, RefCOCO \cite{yu2016modeling}, RefCOCO+ \cite{yu2016modeling}, RefCOCOg \cite{nagaraja2016modeling}), generalized RES dataset gRefCOCO \cite{liu2023gres} and our RefCOCOm benchmark for multi-granularity RES.

\subsection{Datasets}

\textbf{RefCOCO} \cite{yu2016modeling} is one of the largest and most widely used grounding datasets for referring expression segmentation, derived from MSCOCO \cite{lin2014microsoft}. It consists of 142,209 annotated expressions with an average length of 3.6 words, encompassing 50,000 labeled objects across 19,994 images. RefCOCO dataset is divided into 120,624 training samples, 10,834 validation samples, and two test subsets (\ie, test A and test B) comprising 5,657 and 5,095 instances.

\textbf{RefCOCO+} \cite{yu2016modeling} comprises 141,564 natural language expressions, each averaging 3.5 words, and 49,856 target objects across 19,992 images. This dataset adopts a partitioning scheme similar to RefCOCO, including 120,624 training samples, 10,758 validation samples, and two test subsets test A and test B with 5,726 and 4,889 instances respectively. A distinguishing feature of RefCOCO+ is the exclusion of expressions that utilize absolute location terms, thereby presenting an enhanced challenge for the traditional RES task.

\textbf{RefCOCOg} \cite{nagaraja2016modeling}, as the third employed benchmark dataset, features 104,560 referring expressions with a notably longer average length of 8.4 words, describing 54,822 objects across 26,711 images. Unlike the above two datasets, the language expressions in RefCOCOg are collected via Amazon Mechanical Turk, marking a unique aspect to its data collection process. Consistent with prior studies, we use the UMD partitioning standard \cite{hu2016segmentation} for evaluations in this work.

\textbf{gRefCOCO} \cite{liu2023gres} is a more challenging and comprehensive dataset specifically constructed for the GRES task. Unique features of gRefCOCO include the incorporation of negative samples and multi-target expressions, which introduce greater complexity to the classic RES task.
It comprises 278,232 language expressions, among which 80,022 are multi-target expressions (including single-target expressions inherited from RefCOCO \cite{yu2016modeling}) and 32,202 are no-target expressions. These expressions refer to 60,287 distinct instances across 19,994 images, with masks and bounding boxes provided for all target instances. The annotation procedure of gRefCOCO follows ReferIt \cite{kazemzadeh2014referitgame} dataset, maintaining consistency and reliability. Additionally, the data splitting adheres to the UNC partition standard of RefCOCO \cite{hu2016segmentation}, dividing the dataset into a training set, validation set, testA set and testB set.

\subsection{Implementation Details}

\noindent \textbf{Experimental Setup.} 
Our work is implemented based on PyTorch \cite{paszke2019pytorch} and trained on 8 NVIDIA A800 GPUs. We employ Vicuna \cite{chiang2023vicuna} with 7 billion parameters as the LLM, and Vicuna can be seamlessly replaced with more powerful LLMs to further enhance the overall framework's performance.
Considering the critical aspects of scalability and ease of implementation, the vision encoders integrated into our UniRES++ are kept frozen, and LORA-8 is utilized to efficiently adapt LLM during both pre-training and fine-tuning phases. For effective pre-training our framework, we adhere to the same strategy as detailed in \cite{rasheed2024glamm}.
During the joint fine-tuning on grounding tasks at various granularity levels and the region captioning task, 
with a batch size of 256 and approximately 2000 steps per epoch, we adopt the AdamW optimizer to train our UniRES++ for 3 epochs. 
With a linear warm-up strategy for 100 steps during task-oriented fine-tuning, the initial learning rate is set to 5e-4 with a cosine decay schedule.
To be mentioned, for the GRES task, when the number of predicted foreground pixels is less than 50, we consider that the current sample does not contain the corresponding target object. For training objective, we set $\lambda_{lm} = 1.0$, $\lambda_{mask} = 1.0$, $\lambda_{bce} = 2.0$ and $\lambda_{dice} = 0.5$, respectively.

\begin{table*}[htbp]
    \small
    \setlength{\belowcaptionskip}{1.0pt}
    \begin{center}
    \caption{Comparison with previous SOTA methods on our RefCOCOm benchmark in terms of mIoU. Part and Object \& Part denote part-only and multi-grained evaluation settings of our MRES task.}
    \label{tab:results_mres}
    \centering
    \setlength{\tabcolsep}{3.2mm}{\begin{tabular}{l|cc|cc|cc}
    \specialrule{.1em}{.05em}{.05em} 
    \multirow{2}{*}{Methods} & \multicolumn{2}{c|}{val} & \multicolumn{2}{c|}{testA} & \multicolumn{2}{c}{testB} \\
    & Part & Object \& Part  & Part & Object \& Part & Part & Object \& Part  \\
    \midrule
    \multicolumn{7}{l}{\textit{Specialists}} \\
    \midrule
    SeqTR~\cite{zhu2022seqtr} & 13.9 & 28.2 & 12.1 & 22.8 & 18.1 & 34.7 \\
    CRIS~\cite{wang2022cris} & 10.6 & 25.4 & 10.1 & 21.2 & 12.9 & 30.0 \\
    LAVT~\cite{yang2022lavt} & 15.3  & 29.9 & 13.2 & 24.4 & 18.7 & 35.5 \\
    \midrule
    \multicolumn{7}{l}{\textit{Generalists}} \\
    \midrule
    X-Decoder~\cite{zou2023generalized} &  16.2 & 29.5 &  13.6 & 23.6 & 20.3 & 33.8 \\
    SEEM~\cite{zou2023segment} & 16.1 & 29.4 & 13.6 & 23.4 & 20.4 & 33.9 \\
    UniRES \cite{wang2024unveiling} & 19.6 & 34.3 & 16.4 & 27.8 & 25.2 & 41.7 \\
    \midrule
    \rowcolor{mygray} \textbf{UniRES++} & \textbf{27.7} & \textbf{40.8} & \textbf{20.3} & \textbf{35.6} & \textbf{35.3} & \textbf{45.6}\\
    \specialrule{.1em}{.05em}{.05em} 
    \end{tabular}
    }
    \end{center}
    \vspace{-5mm}
\end{table*}

\begin{table*}[htbp]
    \small
    \setlength{\belowcaptionskip}{1.0pt}
    \begin{center}
    \caption{Comparison with the SOTA methods on gRefCOCO. Generalist Methods denotes methods that based on large language model (LLM) and trained on multiple datasets. `FT' denotes that the model is fine-tuned on gRefCOCO. Specialist Methods denote methods designed for RES/GRES and trained exclusively on gRefCOCO. `-' denotes that the result is not provided. }
    \label{tab:results_gres}
    \setlength{\tabcolsep}{2.8mm}{
    \begin{tabular}{l|ccc|ccc|ccc}
    \specialrule{.1em}{.05em}{.05em}
    \multirow{2}{*}{Methods}  & \multicolumn{3}{c|}{val} & \multicolumn{3}{c|}{testA}  & \multicolumn{3}{c}{testB} \\
    & cIoU & gIoU & N-acc. & cIoU & gIoU & N-acc. & cIoU & gIoU & N-acc. \\
    \midrule
    \multicolumn{10}{l}{\textit{Specialists}} \\
    \midrule
    MattNet~\cite{yu2018mattnet} & 47.5 & 48.2 & 41.2 & 58.7 & 59.3 & 44.0 & 45.3 & 46.1 & 41.3 \\
    LTS~\cite{jing2021locate} & 52.3 & 52.7 & - & 61.9 & 62.6 & - & 50.0 & 50.4 & - \\
    VLT~\cite{ding2021vision} & 52.5 & 52.0 & 47.2 & 62.2 & 63.2 & 48.7 & 50.5 & 50.9 & 47.8 \\
    CRIS~\cite{wang2022cris} & 55.3 & 56.3 & - & 63.8 & 63.4 & - & 51.0 & 51.8  & - \\
    LAVT~\cite{yang2022lavt} & 57.6 & 58.4 & 49.3 & 65.3 & 65.9 & 49.3 & 55.0 & 55.8 & 48.5 \\
    ReLA~\cite{liu2023gres} & 62.4 & 63.6 & 56.4 & 69.3 & 70.0 & 59.0 & 59.9 & 61.0 & 58.4 \\
    \midrule
    \multicolumn{10}{l}{\textit{Generalists}} \\
    \midrule
    LISA~\cite{lai2024lisa} & 38.7 & 32.2 & 2.7 & 52.6 & 48.5 & 6.4 & 44.8 & 39.7 & 5.0 \\
    LISA-FT~\cite{lai2024lisa} & 61.8 & 61.6 & 54.7 & 68.5 & 66.3 & 50.0 & 60.6 & 58.8 & 51.9 \\
    GSVA~\cite{xia2024gsva} & 61.7 & 63.3 & 56.5 & 69.2 & 70.1 & 63.5 & 60.3 & 61.3 & 58.4  \\
    GSVA-FT~\cite{xia2024gsva} & 63.3 & 66.5 & 62.4 & 69.9 & 71.1 & 65.3 & 60.5 & 62.2 & 60.6 \\
    \midrule
    \rowcolor{mygray} \textbf{UniRES++} & \textbf{69.9} & \textbf{74.4} & \textbf{74.5} & \textbf{74.5} & \textbf{76.0} & \textbf{70.9} & \textbf{66.6} & \textbf{69.8} & \textbf{70.4} \\
    \specialrule{.1em}{.05em}{.05em}
    \end{tabular}
    }
    \end{center}
    \vspace{-3mm}
\end{table*}

\begin{table*}[htbp]
    \small
    \setlength{\belowcaptionskip}{1.0pt}
    \begin{center}
    \caption{Comparisons with the state-of-the-art approaches on previous three classic RES benchmark datasets in terms of both oIoU and mIoU.
    ``-'' denotes that the result is not provided.}
    \label{tab:results_res}
    \setlength{\tabcolsep}{3.0mm}{
    \begin{tabular}{l|c|ccc|ccc|cc}
    \specialrule{.1em}{.05em}{.05em}
        \multicolumn{2}{c|}{\multirow{2}{*}{Methods}} &  \multicolumn{3}{c|}{RefCOCO} & \multicolumn{3}{c|}{RefCOCO+} & \multicolumn{2}{c}{RefCOCOg} \\
        \multicolumn{2}{c|}{} & val & testA & testB & val & testA & testB & val & test \\
        \midrule
        \multirow{9}{*}{\rotatebox{90}{oIoU}} & EFNet \cite{feng2021encoder} & 62.8 & 65.7 & 59.7 & 51.5 & 55.2 & 43.0 & - & - \\
        & LTS \cite{jing2021locate} & 65.4 & 67.8 & 63.1 & 54.2 & 58.3 & 48.0 & 54.4 & 54.3 \\
        & ReSTR \cite{kim2022restr} & 67.2 & 69.3 & 64.5 & 55.8 & 60.4 & 48.3 & - & - \\
        & ReLA~\cite{liu2023gres} & 73.8 & 76.5 & 70.2 & 66.0 & 71.0 & 57.7 & 65.0 & 66.0 \\
        & X-Decoder~\cite{zou2023generalized} & - & - & - & - & - & - & 64.6 & - \\
        & SEEM~\cite{zou2023segment} & - & - & - & - & - & - & 65.7 & -   \\
        & LISA~\cite{lai2024lisa} & 74.9 & 79.1 & 72.3 & 65.1 & 70.8 & 58.1 & 67.9 & 70.6  \\
        & UniRES \cite{wang2024unveiling} & 77.4 & 80.9 & 74.7 & 69.4 & \textbf{76.1} & 61.4 & 69.0 & 71.7 \\
        \cline{2-10}
        \rowcolor{mygray} \cellcolor{white} & \textbf{UniRES++} & \textbf{80.2} & \textbf{81.8} & \textbf{75.8} & \textbf{71.6} & 76.0 & \textbf{64.4} & \textbf{73.8} & \textbf{74.1} \\
        \midrule
        \multirow{8}{*}{\rotatebox{90}{mIoU}} & VLT \cite{ding2021vision} & 65.7 & 68.3 & 62.7 & 55.5 & 59.2 & 49.4 & 53.0 & 56.7 \\
        & RefTr \cite{li2021referring} & 74.3 & 76.8 & 70.9 & 66.8 & 70.6 & 59.4 & 66.6 & 67.4 \\
        & SeqTR \cite{zhu2022seqtr} & 71.7 & 73.3 & 69.8 & 63.0 & 66.7 & 59.0 & 65.0 & 65.7 \\
        & CRIS \cite{wang2022cris} & 70.5 & 73.2 & 66.1 & 62.3 & 68.1 & 53.7 & 59.9 & 60.4 \\
        & LAVT \cite{yang2022lavt}  & 74.5 & 76.9 & 70.9 & 65.8 & 71.0 & 59.2 & 63.3 & 63.6 \\
        & CM-MaskSD \cite{wang2024cm} & 74.9 & 77.5 & 71.3 & 67.5 & 71.8 & 59.9 & 66.5 & 66.6 \\
        \cline{2-10}
        \rowcolor{mygray} \cellcolor{white} & \textbf{UniRES++} & \textbf{80.8} & \textbf{82.4} & \textbf{77.0} & \textbf{73.6} & \textbf{77.5} & \textbf{67.2} & \textbf{74.4} & \textbf{75.1} \\
        \specialrule{.1em}{.05em}{.05em}
    \end{tabular}
    }
    \end{center}
    \vspace{-5mm}
\end{table*}

\noindent \textbf{Evaluation Metrics.} 
For the classical RES datasets RefCOCO, RefCOCO+, and RefCOCOg, we follow previous works by employing mean intersection over union (mIoU) and overall intersection over union (oIoU) as evaluation metrics. The mIoU quantifies the ratio of the intersection area to the union area between the predicted masks and the ground truth labels across individual test samples, whereas the oIoU assesses the total intersection area over total union area across the entire set of test samples.
For the GRES benchmark dataset gRefCOCO, we follow the benchmark's standard settings by utilizing cumulative intersection over union (cIoU), generalized intersection over union (gIoU), and no-target accuracy (N-acc.) to assess model performance. Specifically, cIoU, which is identical to oIoU, measures the total intersection pixels divided by total union pixels across all samples. The gIoU calculates the average IoU across all samples within each image, providing a nuanced view of model accuracy. Additionally, N-acc. represents the proportion of correctly predicted negative samples, evaluating the model's ability to accurately identify instances with no targets. This comprehensive set of metrics ensures a robust evaluation of model performance across both traditional and generalized RES tasks.

\subsection{Main Results}

\subsubsection{Finer-Grained Part-Level MRES Task}

To evaluate the multi-granularity, particularly part-level grounding performance of our UniRES++ and previous RES SOTA methods for our MRES task, we first conduct an experimental comparison on our newly built RefCOCOm benchmark. 
As presented in Table \ref{tab:results_mres}, we incorporate classic RES methods, including four specialist models (\ie, SeqTR \cite{zhu2022seqtr}, CRIS \cite{wang2022cris}, LAVT \cite{yang2022lavt}), two generalist models (\eg, X-Decoder \cite{zou2023generalized} and SEEM \cite{zou2023segment}), and the previous baseline UniRES \cite{wang2024unveiling} for MRES. 
For fair comparisons, we re-implement these SOTA methods and report their performance on our RefCOCOm benchmark. 
It is evident that both the specialist models and the powerful generalist models perform poorly on RefCOCOm, which requires critical skills in both part-level and object-level referring segmentation.
Benefiting from the MRES-32M dataset, the initial version of the MRES baseline UniRES can better master part-level RES skills and handle the multi-granularity RES task, achieving considerably higher segmentation accuracy. 
In contrast, building upon the MRES-32M dataset as part of the training data, our UniRES++ further enhances performance by introducing grounding data of other target granularities to assist fine-grained V-L understanding and designing multi-granularity feature interactions tailored for unified multi-granularity grounding tasks. 
This results in outstanding performance improvements compared to UniRES.
Ultimately, our UniRES++ establishes new SOTA results for part-level grounding on the three subsets of the RefCOCOm benchmark, achieving scores of 27.7, 20.3, and 35.3, respectively. 
Due to the significantly higher difficulty of part-level RES compared to classic RES, the absolute values of segmentation accuracy are correspondingly lower.
This further emphasizes the importance of researching finer-grained part grounding, where previous SOTA methods have fallen short.

\subsubsection{Generalized Multi-Object GRES Task}

To validate the superiority of our proposed UniRES++ on the multi-object grounding (\ie, GRES) task, we conduct a fair comparison with previous SOTA methods on the gRefCOCO dataset. 
Specifically, we include the most recently proposed LLM-based generalist methods (\eg, LISA \cite{lai2024lisa}, GSVA \cite{xia2024gsva}) for various V-L understanding tasks and specialist methods (\eg, ReLA \cite{liu2023gres}, LAVT \cite{yang2022lavt}) for conventional RES tasks. 
As presented in Table \ref{tab:results_gres}, compared to previous specialist or generalist RES methods, our UniRES++ achieves new SOTA results across all subsets of gRefCOCO. 
Notably, on the gRefCOCO validation, testA, and testB sets, our model surpasses the previously most capable method GSVA by at least $5\%\sim10\%$ after specific finetuning, which represents a significant performance improvement. 
The enhancement is even more pronounced compared to the recently proposed V-L understanding generalist model LISA. 
The remarkable performance improvements in locating multiple instance targets and identifying no-target samples are attributed, on one hand, to the introduction of part-granularity and single-object granularity data, which synergistically enhance multi-object localization tasks, and on the other hand, to the UniRES++ architecture's effective utilization of multi-granularity visual features, enabling the model to establish fine-grained associations between different granularity V-L information.

\subsubsection{Classic Single-Object RES Task}

Besides, we also conduct experiments on the three benchmark datasets for the single object grounding (\ie, classic RES) task. As presented in Table \ref{tab:results_res}, our method significantly outperforms previous approaches in terms of segmentation accuracy across all benchmark datasets. 
Thanks to the multi-granularity feature acquisition and interaction design introduced for multi-granularity grounding tasks, our UniRES++ achieves a more precise mastery of target-level V-L cross-modal associations through joint learning on mutually reinforcing multi-granularity grounding data. 
Whether compared to previous specialist models (\eg, LAVT \cite{yang2022lavt} and LLM-based LISA \cite{lai2024lisa}) for classic RES or V-L understanding generalist models (\eg, X-Decoder \cite{zou2023generalized} and SEEM \cite{zou2023segment}), our proposed grounding generalist UniRES++ demonstrates absolute advantages in both oIoU and mIoU metrics. 
Furthermore, compared to the recently proposed SOTA methods, our UniRES++ greatly surpasses previous methods by nearly $3\%\sim5\%$ on the majority of metrics across the three datasets, fully demonstrating the importance of joint learning on multi-granularity grounding data and the multi-granularity feature interaction design for RES tasks.

\vspace{-5pt}
\subsection{Qualitative Analysis}

\vspace{-12pt}
\begin{figure}[htbp]
    \centering
    \includegraphics[width=0.49\textwidth]{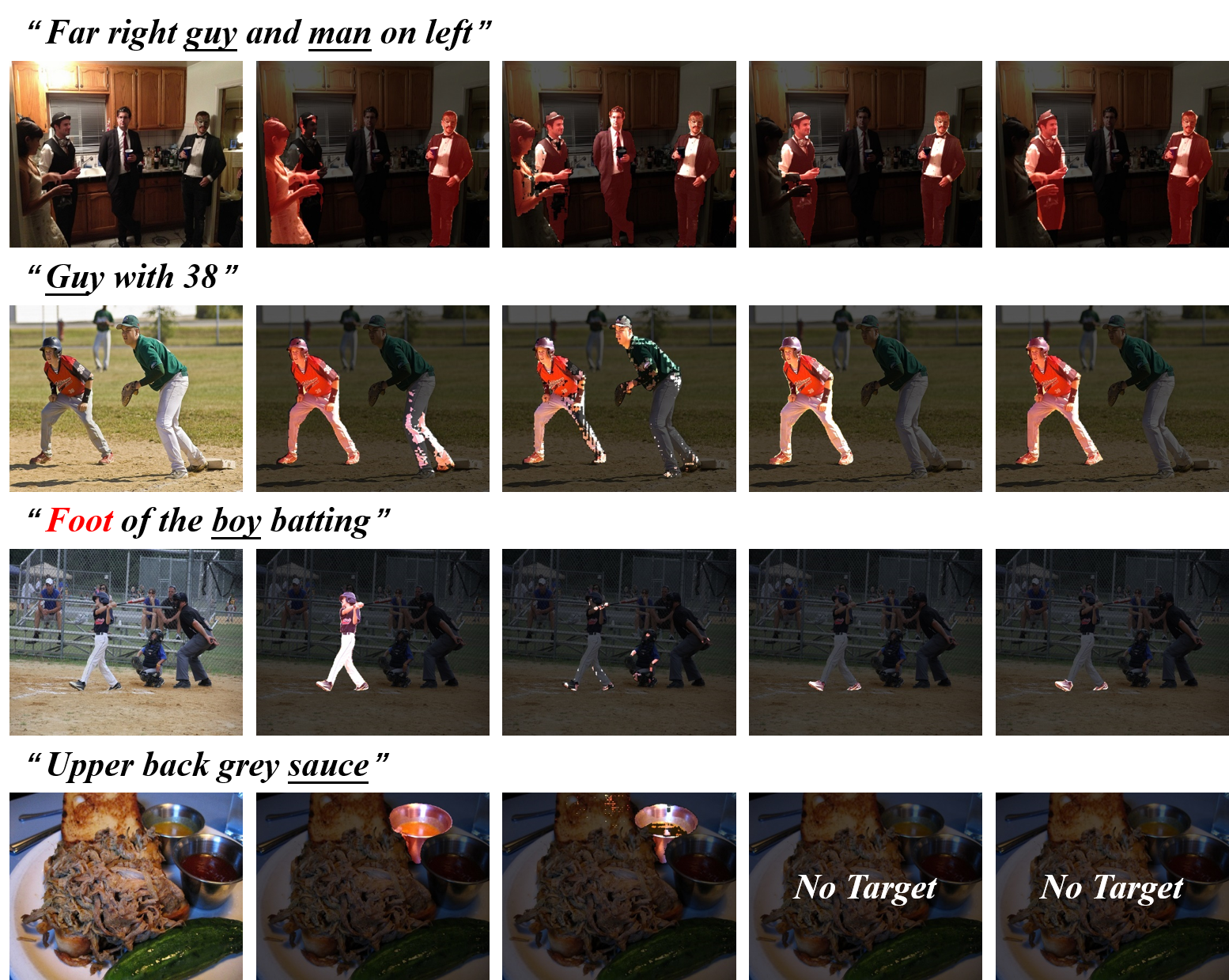}
    \begin{tabu} to 1.0\linewidth{X[1.0c] X[1.0c] X[1.0c] X[1.0c] X[1.0c]} 
        \scriptsize{(a) Image} &  \scriptsize{(b) ReLA} &  \scriptsize{(c) LISA}  &  \scriptsize{\textbf{(e) Ours}} &  \scriptsize{(d) GT} \\
    \end{tabu}
    \vspace{-15pt}
    \caption{The visual comparison of segmentation results on both RefCOCOm and gRefCOCO validation set across different. (a) the input image. (b) ReLA. (c) LISA. (d) our UniRES++. (e) the ground truth (GT). }
    \label{fig_SOTAComparison_vis_1}
    \vspace{-8pt}
\end{figure}

\subsubsection{Visual Comparison with SOTA Methods}

\begin{figure}[htbp]
    \centering
    \includegraphics[width=0.49\textwidth]{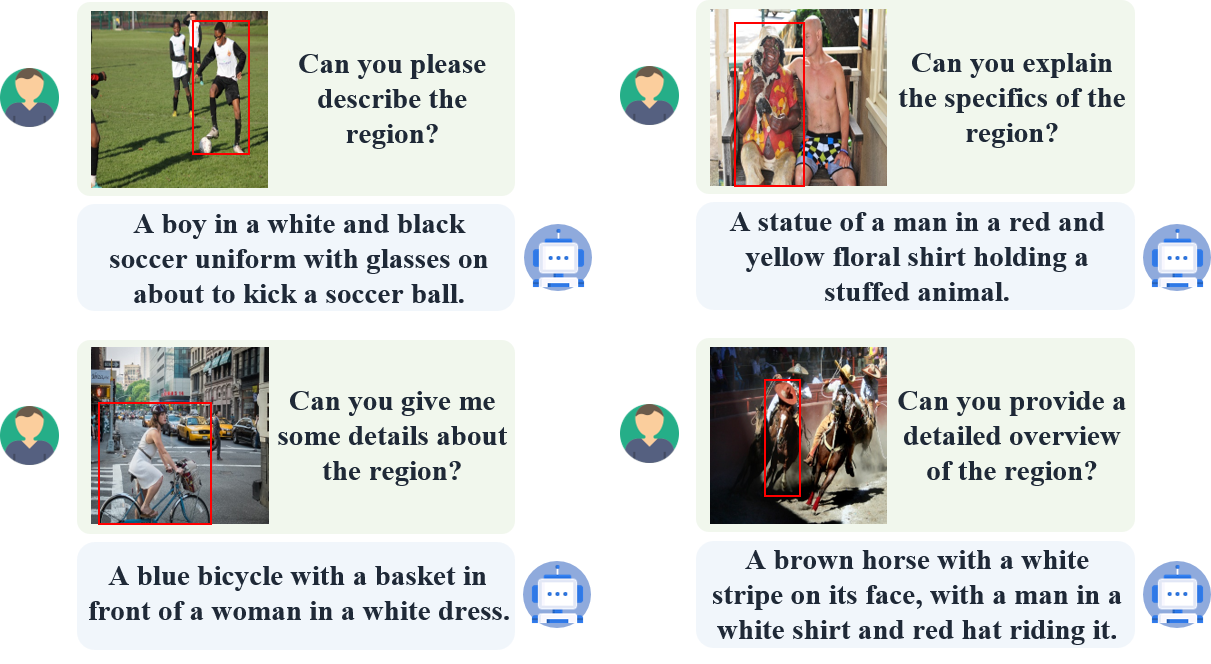}
    \vspace{-20pt}
    \caption{Qualitative presentation of our grounding generalist UniRES++’s performance on region-level dense captioning task.}
    \label{fig_SOTAComparison_vis_2}
    \vspace{-5pt}
\end{figure}

To validate the segmentation quality of our framework, we first adopt previous SOTA specialist methods for classic RES and GRES tasks, including ReLA \cite{liu2023gres}, LISA \cite{lai2024lisa} and our grounding generalist UniRES++, for qualitative comparison. 
As shown in Fig. \ref{fig_SOTAComparison_vis_1}, we present a comparison across four target granularity localization tasks, including one expression referring to multi-object, single-object, single-part, and no-target scenarios, from the first to the fourth row, respectively. 
The results in Fig. \ref{fig_SOTAComparison_vis_1} demonstrate that our UniRES++ surpasses previous specialists by realizing both finer-grained V-L understanding and more unified multi-granularity grounding.
From the first and second rows, it can be observed that although previous specialist methods perform well in single and multiple object localization tasks, our generalist generates much better segmentation masks of the target objects, greatly reducing over-segmentation and under-segmentation errors. 
In contrast, the results in the third row convince that when addressing the more challenging part-level grounding task, our UniRES++ can locate and segment the referring target regions more accurately, whereas the other specialist methods fail to achieve the same level of performance.
Besides, thanks to the accurate modeling of fine-grained V-L feature associations, our UniRES++ can also well handle other V-L understanding tasks beyond grounding, such as fine-grained region captioning, as shown in Fig. \ref{fig_SOTAComparison_vis_2}.

\vspace{-5pt}

\subsubsection{Grounding Capabilities In Complex Scenes}

\begin{figure*}[htbp]
    \centering
    \includegraphics[width=1.0\textwidth]{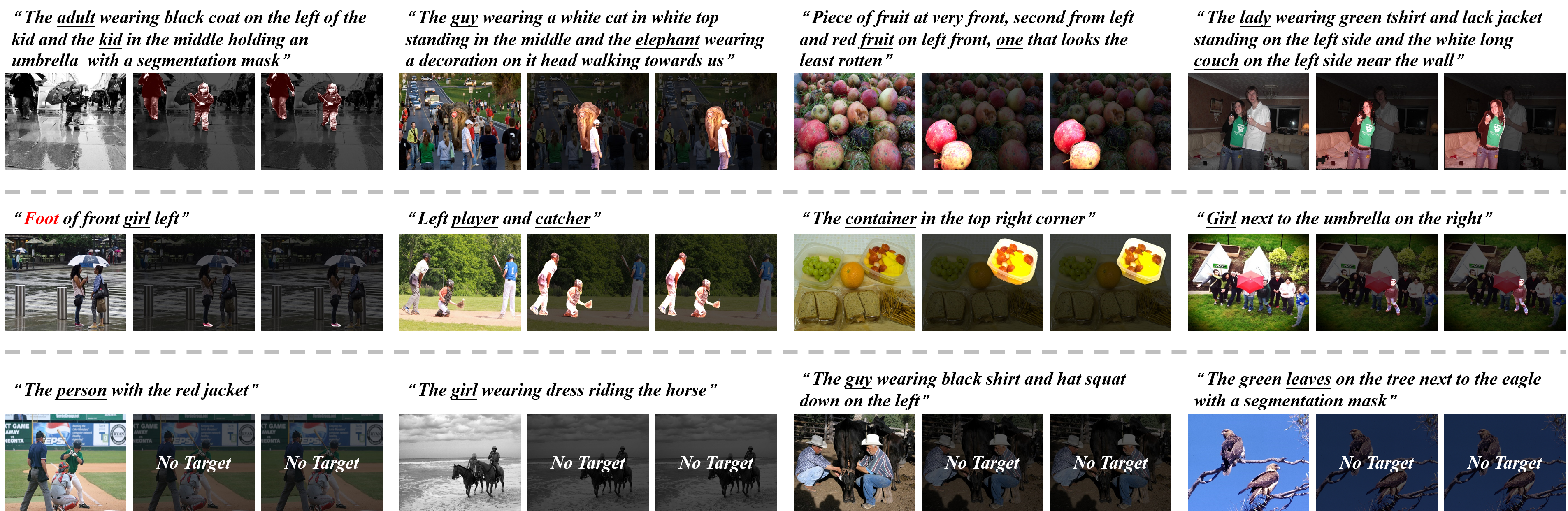}
    \begin{tabu} to 1.0\linewidth{X[1.0c] X[1.0c] X[1.0c] X[1.0c] X[1.0c] X[1.0c] X[1.0c] X[1.0c] X[1.0c] X[1.0c] X[1.0c] X[1.0c]} 
        \scriptsize{(a) Image} &  \scriptsize{(b) Ours} &  \scriptsize{(c) GT}  &  \scriptsize{(a) Image} &  \scriptsize{(b) Ours} &  \scriptsize{(c) GT} & \scriptsize{(a) Image} &  \scriptsize{(b) Ours} &  \scriptsize{(c) GT} & \scriptsize{(a) Image} &  \scriptsize{(b) Ours} &  \scriptsize{(c) GT} \\
    \end{tabu}
    \vspace{-15pt}
    \caption{{Qualitative presentation of our grounding generalist's performance in complex scenarios, including referring expressions that are much longer or contain spatial relationships, and the given descriptions may not correspond to any target. (a) the input image. (b) our UniRES++. (c) the ground truth (GT).}}
    \label{fig_more_visualization}
    \vspace{-15pt}
\end{figure*}

Additionally, to demonstrate the potential of our grounding generalist UniRES++ in mastering multi-granularity target localization, we visualize the grounding results of UniRES++ in relatively complex scenarios, including referring expressions that are much longer or contain complex spatial relationships, as well as real-world scenes where the given text descriptions by users may not correspond to any target. 
These three complex scenarios correspond to the first to third rows from top to bottom in Fig. \ref{fig_more_visualization}. 
As shown in the first and second rows in Fig. \ref{fig_more_visualization}, thanks to our multi-granularity grounding data and the powerful text understanding capabilities of the large language models, our UniRES++ can effectively and accurately localize the corresponding targets with the given long texts and referring instructions with spatial relationships. 
Moreover, it is clear from the third row that false grounding cases are effectively and accurately handled by our UniRES++ model (\ie, it does not output any grounding mask for targets that do not exist as referred to by the current descriptions).

\subsubsection{Temporary Failure Case}

\begin{figure}[htbp]
    \centering
    \includegraphics[width=0.49\textwidth]{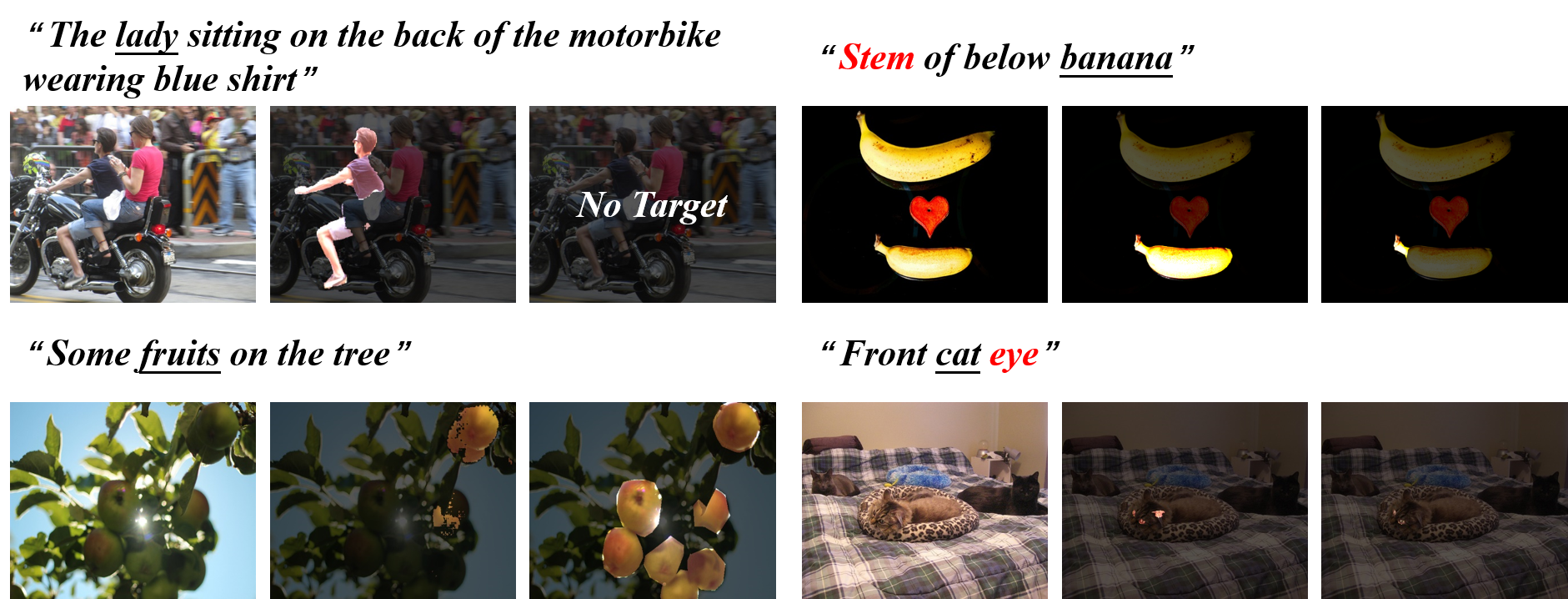}
    \begin{tabu} to 1.0\linewidth{X[1.0c] X[1.0c] X[1.0c] X[1.0c] X[1.0c] X[1.0c]} 
        \scriptsize{(a) Image} &  \scriptsize{(b) Ours} &  \scriptsize{(c) GT}  &  \scriptsize{(a) Image} &  \scriptsize{(b) Ours} &  \scriptsize{(c) GT}  \\
    \end{tabu}
    \vspace{-20pt}
    \caption{{Failure case presentation of our grounding generalist UniRES++’s performance on generalized RES (GRES) and multi-granularity RES (MRES) tasks. (a) the input image. (b) our UniRES++. (c) the ground truth (GT).}}
    \label{fig_failure_case}
    \vspace{-10pt}
\end{figure}

Moreover, to explore the current limitations of our UniRES++'s grounding capability, we have visualized some failure cases of UniRES++. As shown in Fig. \ref{fig_failure_case}, the failure cases of our UniRES++ can be mainly divided into two categories: 1) For the highly challenging MRES task, UniRES++ still has some incorrect grounding results for part regions that have varying shapes and small areas (as Fig. \ref{fig_failure_case} second column); 2) For the more practical GRES task, in grounding cases where a single text description may refer to multiple corresponding targets, UniRES++ sometimes still produces incomplete grounding results (as the bottom left part of Fig. \ref{fig_failure_case}), and in grounding cases where a single text description may refer to no corresponding target, occasional false positive grounding errors still occur (as the top left part of Fig. \ref{fig_failure_case}). Both of these two issues need to be addressed by scaling up the model structure and collecting more high-quality grounding data to directly enhance the model's grounding capabilities and robustness.

\subsection{Ablation Studies}

To investigate the advantages of our proposed method across existing grounding tasks at different granularity levels, we conduct comprehensive ablation experiments on the validation sets of both our RefCOCOm and gRefCOCO datasets. The tables below involve three evaluation settings for our RefCOCOm dataset: either taking only objects or only parts as RES targets, and a mixed setting that combines both granularity levels.

\begin{table}[htbp]
    \small
    \setlength{\belowcaptionskip}{1.0pt}
    \begin{center}
    \caption{Ablation study on the input granularity and high resolution of the multi-grained vision flow in our UniRES++. Object and Part denote the introduction of auxiliary object-level and part-level feature representations.}
    \vspace{-8pt}
     \setlength{\tabcolsep}{0.5mm}{\begin{tabular}{c|c|c|ccc|ccc}
      \specialrule{.1em}{.05em}{.05em} 
      \multirow{2}{*}{Object} & \multirow{2}{*}{Part} & \multirow{2}{*}{High-Res} & \multicolumn{3}{c|}{RefCOCOm-val} & \multicolumn{3}{c}{gRefCOCO-val} \\
       &  &  & Object & Part & Object\!\;\&\;\!Part & cIoU & gIoU & N-acc. \\
        \hline
        &  & - & 78.1 & 25.7 & 38.6 & 68.4 & 70.8 & 67.5 \\
        \ding{51} &  & 1024 & 80.7 & 25.8 & 39.3 & \textbf{70.1} & 74.1 & 74.2 \\ 
        & \ding{51} & 1024 & 77.8 & 27.5 & 39.9 & 68.3 & 71.0 & 70.9 \\ 
        \rowcolor{mygray} \ding{51} & \ding{51} & 1024 & \textbf{80.8} & \textbf{27.7} & \textbf{40.8} & 69.9 & \textbf{74.4} & \textbf{74.5} \\  
        \hline
        \ding{51} & \ding{51} & 512 & 80.5 & 25.2 & 38.9 & 69.7 & 72.5 & 69.2
        \\ 
        \ding{51} & \ding{51} & 768 & 80.6 & 25.8 & 39.3 & 69.5 & 72.9 & 70.6 \\ 
        \rowcolor{mygray} \ding{51} & \ding{51} & 1024 & \textbf{80.8} & \textbf{27.7} & \textbf{40.8} & \textbf{69.9} & \textbf{74.4} & \textbf{74.5} \\  
        \ding{51} & \ding{51} & 2048 & 80.6 & 27.0 & 40.2 & 69.7 & 72.7 & 70.0 \\ 
    \specialrule{.1em}{.05em}{.05em}
    \end{tabular}
    \vspace{-6mm}
    \label{tab:ablation_data_granularity}}
    \end{center}
\end{table}

\noindent \textbf{Multi-Granularity Vision Flow.} 
We first investigate the impact of input data granularity and the high resolution of fine-grained feature sources in our multi-granularity vision flow design within UniRES++ model. 
As shown in the upper part of Table \ref{tab:ablation_data_granularity}, thanks to the LLM's exceptional sequence modeling capabilities, the introduction of additional object-level or part-level detailed features enhances the model's accuracy in grounding tasks at the corresponding granularity levels. 
Despite that our UniRES++ has already achieved excellent results across various RES tasks at the three target granularities by capturing only image-level visual features, the object and part region features serve as supplementary input representations, collaboratively enabling the model to more precisely comprehend fine-grained information. 
Furthermore, the results in the lower part of Table \ref{tab:ablation_data_granularity} show that appropriately increasing the resolution of the auxiliary object/part-level feature sources (\ie, high-resolution images $I_h$) steadily improves the model's performance by providing richer fine-grained details. A resolution of 1024 represents a sweet spot for the model's grounding capability. When the high resolution is further increased to 2048, the model's performance begins to decline. We believe this is due to the excessively high resolution deviating too much from the encoder's original resolution during training, thereby compromising the representation quality of the object and part detailed features.

\begin{table}[htbp]
    \small
    \setlength{\belowcaptionskip}{1.0pt}
    \begin{center}
    \caption{Ablation study on the introduced fine-tuning data scale.}
     \vspace{-1mm}
     \setlength{\tabcolsep}{1.6mm}{\begin{tabular}{c|ccc|ccc}
      \specialrule{.1em}{.05em}{.05em} 
      \multirow{2}{*}{Ratios} & \multicolumn{3}{c|}{RefCOCOm-val} & \multicolumn{3}{c}{gRefCOCO-val} \\
       & Object & Part & Object\!\;\&\;\!Part & cIoU & gIoU & N-acc. \\
        \hline
        25\%  &  78.8  &  9.2  &  26.4  &  66.1  &  70.6   & 68.3  \\ 
        50\%  &  80.1  &  18.0  &  33.3  &  67.8  & 72.2 & 70.3  \\ 
        75\%  &  80.2  &  25.2  &  38.8  &  68.8  &  72.8   & 70.3  \\ 
        \rowcolor{mygray} 100\% & \textbf{80.8} & \textbf{27.7} & \textbf{40.8} & \textbf{69.9} & \textbf{74.4} & \textbf{74.5} \\ 
    \specialrule{.1em}{.05em}{.05em}
    \end{tabular}
    \vspace{-15pt}
    \label{tab:ablation_data_scale}}
    \end{center}
\end{table}

\noindent \textbf{Data Scaling Properties.}
Next, we explore the effect of different percentages of training samples during task-oriented fine-tuning. As mentioned previously, the training data at this stage comprises the existing grounding data at all granularity levels and the region captioning task data that aids in better fine-grained understanding. The results are presented in Table \ref{tab:ablation_data_scale}. It is evident from Table \ref{tab:ablation_data_scale} that the model performances for the included grounding tasks at various granularities consistently improve as more training samples are introduced. As the proportions of introduced training samples continue to increase, there are no signs of diminishing returns in model performance, suggesting that our UniRES++ framework has significant potential when scaling up the training data.

\vspace{-5pt}
\begin{table}[htbp]
    \small
    \setlength{\belowcaptionskip}{1.0pt}
    \begin{center}
    \caption{Ablation study on the structural designing for the multi-granularity feature exploitation in UniRES++. DS, AGI and DRW represent the design of decoupling [SEG] token for object/part granularities, adjacent granularity interaction and granularity re-weighting in decoder part.
    }
     \setlength{\tabcolsep}{0.6mm}{\begin{tabular}{c|c|c|ccc|ccc}
      \specialrule{.1em}{.05em}{.05em} 
      \multirow{2}{*}{DS} & \multirow{2}{*}{AGI} & \multirow{2}{*}{DRW} & \multicolumn{3}{c|}{RefCOCOm-val} & \multicolumn{3}{c}{gRefCOCO-val} \\
       &  &  &  Object & Part & Object\!\;\&\;\!Part & cIoU & gIoU & N-acc. \\
        \hline
        &  &  & 80.2 & 25.6 & 39.1 & 69.2 & 71.7 & 67.8 \\ 
        \ding{51} &  &  & 80.5 & 26.2 & 39.6 & 69.6 & 72.5 & 68.5 \\ 
        \ding{51} & \ding{51} &  & 80.6 & 27.0 & 40.2 & \textbf{70.1} & 73.4 & 71.8 \\ 
        \rowcolor{mygray} \ding{51} & \ding{51} & \ding{51} & \textbf{80.8} & \textbf{27.7} & \textbf{40.8} & 69.9 & \textbf{74.4} & \textbf{74.5} \\ 
    \specialrule{.1em}{.05em}{.05em}
    \end{tabular}
    \vspace{-4mm}
    \label{tab:ablation_interaction_granu}}
    \end{center}
\end{table}

\noindent \textbf{Multi-Granularity Feature Exploitation.}
Then, we explore the impact of various structural designs for multi-grained visual feature interaction in our UniRES++, including the separation of the special [SEG] token for object and part granularities, adjacent granularity interaction, and granularity re-weighting in the decoder part.
As presented in Table \ref{tab:ablation_interaction_granu}, the hierarchical incorporation of these three components in the multi-granularity feature interaction design of the unified grounding architecture yields significant performance improvements. Moreover, the simultaneous introduction of these three elements synergistically enhances the overall grounding architecture. 
Additionally, within the multi-granularity vision flow, the feature representations fed into the decoding stage are re-weighted using corresponding granularity features that have been strengthened through adjacent granularity guidance, thereby facilitating more accurate fine-grained V-L understanding by our unified grounding model.

\vspace{-5pt}
\begin{table}[htbp]
    \small
    \setlength{\belowcaptionskip}{1.0pt}
    \begin{center}
    \caption{Ablation study on the introduced grounding data for training unified UniRES++. 
    1-to-O, 1-to-P and 1-to-MO separately denote the data of RefCOCO/RefCOCO+/RefCOCOg, MRES-32M and gRefCOCO datasets.}
     \setlength{\tabcolsep}{0.8mm}{\begin{tabular}{ccc|ccc|ccc}
      \specialrule{.1em}{.05em}{.05em}
       \multicolumn{3}{c|}{Grounding Data} & \multicolumn{3}{c|}{RefCOCOm-val} & \multicolumn{3}{c}{gRefCOCO-val} \\
       \multirow{1}{*}{1-to-O} & \multirow{1}{*}{1-to-P} & \multirow{1}{*}{1-to-MO} & Obj & Part & Obj\!\;\&\;\!Part & cIoU & gIoU & N-acc. \\
        \hline
        \ding{51} &  &   & 80.5  & 10.2 & 27.6 & 37.1 & 32.5 & 0.0   \\ 
        \ding{51} & \ding{51}  &   &  80.7 & 25.7 & 39.3  & 37.1  & 32.5  &  0.0  \\ 
        \ding{51} &   & \ding{51} & 80.6  & 10.5 &  27.8  & 69.3  & 72.4  &  69.7  \\ 
        \rowcolor{mygray} \ding{51} & \ding{51} & \ding{51} & \textbf{80.8} & \textbf{27.7} & \textbf{40.8} & \textbf{69.9} & \textbf{74.4} & \textbf{74.5}  
        \\ 
    \specialrule{.1em}{.05em}{.05em}
    \end{tabular}
    \label{tab:ablation_multi_task}}
    \vspace{-4mm}
    \end{center}
\end{table}

\noindent \textbf{Multi-Granularity Complementary Property.}
Besides, we also conduct ablation studies on the synergistic complementarities between multi-granularity grounding tasks. 
As shown in Table \ref{tab:ablation_multi_task}, 1-to-O, 1-to-P, and 1-to-MO respectively denote the grounding data from RefCOCO/RefCOCO+/RefCOCOg (one expression refers to one object region), MRES-32M (one expression refers to one part region), and gRefCOCO (one expression refers to multiple objects) datasets. 
It is clearly observed that, based on introducing single-object grounding data alone, either introducing multi-object grounding data or single-part region localization data significantly enhances the model performance on the corresponding granularity grounding tasks. 
Moreover, it is noteworthy that there exists a synergistic complementarity between different granularity data. 
For example, it's clear from the first and second rows or the first and third rows of the Table \ref{tab:ablation_multi_task}, the introduction of part and multi-object grounding data leads to reciprocal improvements in the model's performance on multi-object and part-level region localization tasks. 
Ultimately, with the combined support and mutual enhancement of grounding data across three target granularities, our unified model achieves the best general grounding performance by establishing dense V-L associations for different granularity targets, marking an important attempt towards more general V-L understanding.

\vspace{-5pt}
\begin{table}[htbp]
    \small
    \setlength{\belowcaptionskip}{1.0pt}
    \begin{center}
    \caption{Ablation study on the effectiveness of our grounding generalist UniRES++ over existing grounding MLLM, GLaMM. GLaMM and GLaMM* represent the original GLaMM model and the model obtained through fine-tuning with the same training recipe as our UniRES++, respectively.}
     \vspace{-1mm}
     \setlength{\tabcolsep}{1.6mm}{\begin{tabular}{c|ccc|ccc}
      \specialrule{.1em}{.05em}{.05em} 
      \multirow{2}{*}{Methods} & \multicolumn{3}{c|}{RefCOCOm-val} & \multicolumn{3}{c}{gRefCOCO-val} \\
       & Object & Part & Obj\!\;\&\;\!Part & cIoU & gIoU & N-acc. \\
        \hline
        GLaMM  &  79.8  &  21.7  &  36.0  &  38.6  &  33.1  & 1.2  \\ 
        GLaMM* &  79.2  &  25.4  &  38.9  &  68.7  &  68.6  & 69.4  \\ 
        \rowcolor{mygray} \textbf{Ours} & \textbf{80.8} & \textbf{27.7} & \textbf{40.8} & \textbf{69.9} & \textbf{74.4} & \textbf{74.5} \\ 
    \specialrule{.1em}{.05em}{.05em}
    \end{tabular}
    \vspace{-5pt}
    \label{tab:data_model_effectiveness}}
    \end{center}
\end{table}

\noindent \textbf{GLaMM V.S. UniRES++.}
At last, to further demonstrate the advantages of UniRES++ as a grounding generalist, we conduct a comprehensive performance comparison with one of the most advanced models, GLaMM \cite{rasheed2024glamm}. 
The quantitative results are shown in Table \ref{tab:data_model_effectiveness}. 
Specifically, the first and third rows in Table \ref{tab:data_model_effectiveness} present the performance of the original GLaMM model and our UniRES++, respectively. 
It can be clearly seen that although GLaMM achieves comparable performance to UniRES++ on single object grounding task, it lags significantly behind UniRES++ on MRES task requiring part-region localization and more practical GRES task.
It is important to note that the original GLaMM's weight corresponding to the first row is pre-trained on the high-quality GranD dataset constructed in \cite{rasheed2024glamm} and then fine-tuned on the classic RES datasets. 
Moreover, to provide the most intuitive and fair comparison to prove that our UniRES++ architecture is more advantageous under the same training recipe, we conduct additional comparative experiments. As shown in the comparison between the second and third rows in Table \ref{tab:data_model_effectiveness}, although GLaMM achieves significant overall performance improvements on the included benchmarks after being trained with our fine-tuning recipe, our UniRES++ still demonstrates an absolute advantage over GLaMM on the existing benchmark datasets (including gRefCOCO for generalized RES, RefCOCO, RefCOCO+, RefCOCOg for classic RES tasks, and our RefCOCOm for multi-granularity RES tasks), especially on fine-grained MRES and practical GRES tasks.

\section{Conclusion, Limitation and Future Work}
\label{Conclusion}

In this work, we move beyond the limitations of prior research, which typically suffer from limited granularity in visual target understanding and constrained fine-grained V-L comprehension, taking a significant step towards finer-grained part-level comprehension and granularity-unified RES.
We propose a novel MRES task and build an evaluation benchmark RefCOCOm based on manual annotation.
Furthermore, to advance visual grounding at both the object and part levels towards finer-grained V-L understanding, we construct the largest grounding dataset to date, MRES-32M, which is also the first to provide part-level V-L annotations.
In addition, we have developed UniRES++, a powerful multi-granularity RES generalist MLLM. 
As the first unified RES framework across various visual granularities, UniRES++ realizes new SOTA performance across existing RES tasks, including classic RES, generalized RES (GRES), and our MRES tasks.
We plan to release our RefCOCOm benchmark, MRES-32M dataset, and UniRES++ model to the community, aiming to facilitate future research in fine-grained visual grounding and inspire new studies in this domain.

One potential limitation of this work is that both the data scale of MRES-32M dataset and the model capacity of UniRES++ could be further scaled up to push SOTA performance even higher.
Additionally, our current focus is primarily on unifying RES tasks at various granularity levels with visual masks as output. 
While our framework showcases new, essential part-level referring segmentation capabilities and demonstrates a comprehensive understanding of visual targets across different granularities, its ability to handle V-L conversation tasks requires further exploration. 
Although UniRES++ can generate textual responses with the integration of large language models, enabling it to perform image captioning and fine-grained region captioning, its capabilities on other V-L tasks such as visual question answering remain underexplored. 
Moreover, our UniRES++ grounding algorithm has not yet been deployed on real-world robotic systems, which will be a focus of future work. 
These all open up an exciting future research direction: the development of a more general and powerful framework based on MLLMs that can interact with user-provided textual and visual inputs across multiple levels of granularity.

{\small
    \bibliographystyle{IEEEtran}
    \bibliography{reference}
}

\end{document}